\documentclass[final,3p,times,twocolumn]{elsarticle}
\usepackage{amssymb}
\usepackage{amsmath}
\usepackage{bm}
\usepackage{tabularx}
\usepackage[table,svgnames,dvipsnames]{xcolor}
\usepackage{wrapfig}
\usepackage{multirow}
\usepackage{graphicx}
\usepackage{subcaption}
\usepackage{booktabs}
\definecolor{blue}{RGB}{33,143,183}
\usepackage[
    bookmarks=true,
    colorlinks=true,
    linkcolor=blue,
    urlcolor=blue,
    citecolor=blue,
    linktocpage=true,   % makes the page number as hyperlink in table of content
    hyperindex=true
    ]{hyperref}
\AtEndPreamble{
    \usepackage{cleveref}
    \crefname{section}{\textcolor{blue}{Sec.}}{\textcolor{blue}{Secs.}}
    \crefname{table}{\textcolor{blue}{Table}}{\textcolor{blue}{Tables}}
    \crefname{figure}{\textcolor{blue}{Fig.}}{\textcolor{blue}{Figs.}} 
    \crefname{appendix}{}{}
}

\journal{XXX}

\begin{document}

\begin{frontmatter}


\title{Object Retrieval for Visual Question Answering with Outside Knowledge}

\author{Shichao Kan$^a$,
        Yuhai Deng$^{b}$,
        Jiale Fu$^{c}$,
        Lihui Cen$^b$,
        Zhe Qu$^a$,
        Linna Zhang$^{d}$,
        Yixiong Liang$^{a}$,
        Yigang Cen$^{e,f}$
} %% Author name

\affiliation{organization={School of Computer Science and Engineering, Central South University},%Department and Organization
            postcode={410083}, 
            city={Changsha},
            state={Hunan},
            country={China}}
\affiliation{organization={School of Automation, Central South University},%Department and Organization
            postcode={410083}, 
            city={Changsha},
            state={Hunan},
            country={China}}
\affiliation{organization={Dundee International Institute, Central South University},%Department and Organization
            postcode={410083}, 
            city={Changsha},
            state={Hunan},
            country={China}}
\affiliation{organization={College of Mechanical Engineering, Guizhou University},%Department and Organization
            postcode={550025}, 
            city={Guiyang},
            state={Guizhou},
            country={China}}
\affiliation{organization={Institute of Information Science, School of Computer and Information Technology, Beijing Jiaotong University},%Department and Organization
            postcode={100044}, 
            city={Beijing},
            country={China}}
\affiliation{organization={Beijing Key Laboratory of Advanced Information Science and Network Technology},%Department and Organization
            postcode={100044}, 
            city={Beijing},
            country={China}}
% \affiliation{organization={Department of Electrical and Electronic Engineering, Southern University of Science and Technology},%Department and Organization
%             city={Shenzhen},
%             country={China}}
% \affiliation{organization={Pengcheng Lab},%Department and Organization
%             postcode={518066},
%             city={Shenzhen},
%             country={China}}

%% Abstract
\begin{abstract}
Retrieval-augmented generation (RAG) with large language models (LLMs) plays a crucial role in question answering, as LLMs possess limited knowledge and are not updated with continuously growing information. Most recent work on RAG has focused primarily on text-based or large-image retrieval, which constrains the broader application of RAG models. We recognize that object-level retrieval is essential for addressing questions that extend beyond image content. To tackle this issue, we propose a task of object retrieval for visual question answering with outside knowledge (OR-OK-VQA), aimed to extend image-based content understanding in conjunction with LLMs. A key challenge in this task is retrieving diverse objects-related images that contribute to answering the questions. To enable accurate and robust general object retrieval, it is necessary to learn embeddings for local objects. This paper introduces a novel unsupervised deep feature embedding technique called multi-scale group collaborative embedding learning (MS-GCEL), developed to learn embeddings for long-tailed objects at different scales. Additionally, we establish an OK-VQA evaluation benchmark using images from the BelgaLogos, Visual Genome, and LVIS datasets. Prior to the OK-VQA evaluation, we construct a benchmark of challenges utilizing objects extracted from the COCO 2017 and VOC 2007 datasets to support the training and evaluation of general object retrieval models. Our evaluations on both general object retrieval and OK-VQA demonstrate the effectiveness of the proposed approach. The code and dataset will be publicly released for future research.
\end{abstract}
% \input{sec/X2_highlights}

%% Keywords

\end{frontmatter}


\section{Introduction}
\label{sec1}
\begin{figure}[t]
    \begin{center}
    \includegraphics[width=1.0\linewidth]{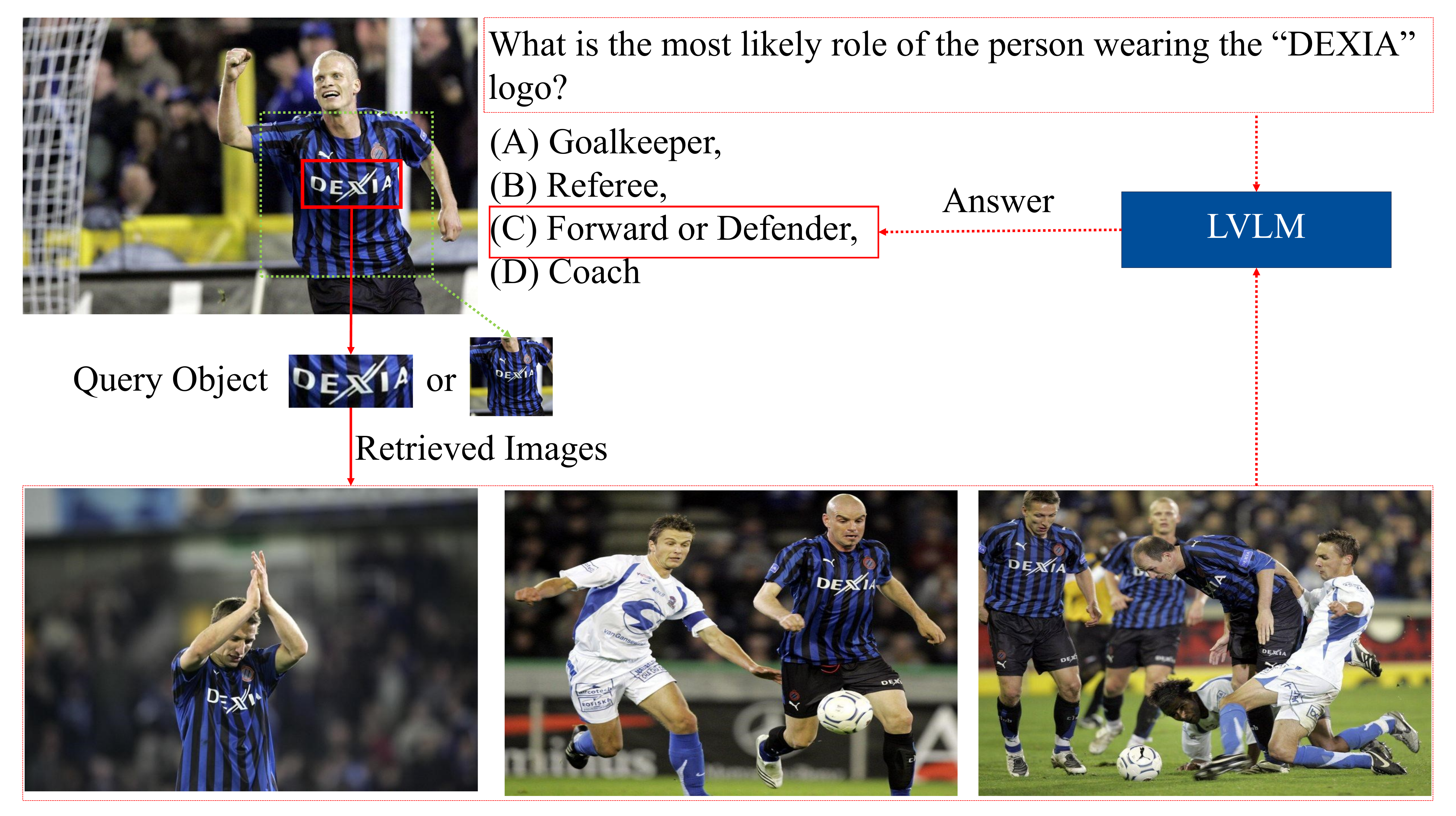}
    \end{center}
    \vspace{-2mm}
	\caption{Object Retrieval for Visual Question Answering with Outside Knowledge (OR-OK-VQA). Images are retrieved from the BelgaLogos dataset~\cite{joly2009logo}.}
	\vspace{-4mm}
	\label{fig:motivate1}
\end{figure}

Visual Question Answering (VQA)~\cite{lin2022retrieval} based on Large Vision-Language Models (LVLMs)~\cite{liu2024visual,wu2024v} has seen significant progress and growing attention. However, VQA becomes particularly challenging when the answer to a question is not explicitly present in the image~\cite{marino2019ok}. To address this, current methods often rely on Retrieval-Augmented Generation (RAG)~\cite{zhao2023retrieving,zhao2024retrieval}, which operates in two phases: retrieval and generation. In the retrieval phase, contextually relevant information is fetched from a large pool of resources, while in the generation phase, LVLMs use this retrieved knowledge to formulate an answer. Given that the quality of retrieval directly impacts the accuracy of LVLMs’ responses, and since many questions pertain to specific objects within an image, it is crucial for retrieval to focus on object-level information rather than the entire image. As illustrated in~\cref{fig:motivate1}, when asked, ``What is the most likely role of the person wearing the `DEXIA' logo?'', it’s difficult to answer based solely on the given image. However, if multiple images related to the `DEXIA' logo are retrieved, the LVLMs can deduce that this person is most likely a forward or defender. This motivates our goal of developing a robust object-centric retrieval method that identifies relevant images at the object level, enhancing precision and relevance for more accurate VQA.

Object retrieval, as explored in existing research~\cite{jiang2015randomized,tolias2016particular-object}, involves locating similar images containing a given object. However, this task is fraught with challenges, as target objects are often embedded in cluttered backgrounds, occupy small or unpredictable portions of an image, and can vary significantly in scale, viewpoint, color, and partial occlusions. While many object retrieval algorithms have been developed for specialized tasks such as person re-identification~\cite{kan2019supervised,zhang2023poar}, vehicle re-identification~\cite{kan2022contrastive}, clothing retrieval~\cite{kan2022local}, and logo retrieval~\cite{jiang2015randomized}, these methods are often trained under independent and identically distributed (\textit{i.i.d.}) assumptions. This limits their generalizability to non-\textit{i.i.d.} scenarios, such as VQA. For example, an algorithm optimized for pedestrian datasets may excel at person retrieval but struggle with vehicle retrieval, making it unsuitable for the broad demands of VQA. To overcome this limitation, our objective is to develop a general object retrieval algorithm capable of handling both seen and unseen objects across diverse tasks, thus broadening its applicability to VQA.

\begin{figure}[t]
    \begin{center}
    \includegraphics[width=1.0\linewidth]{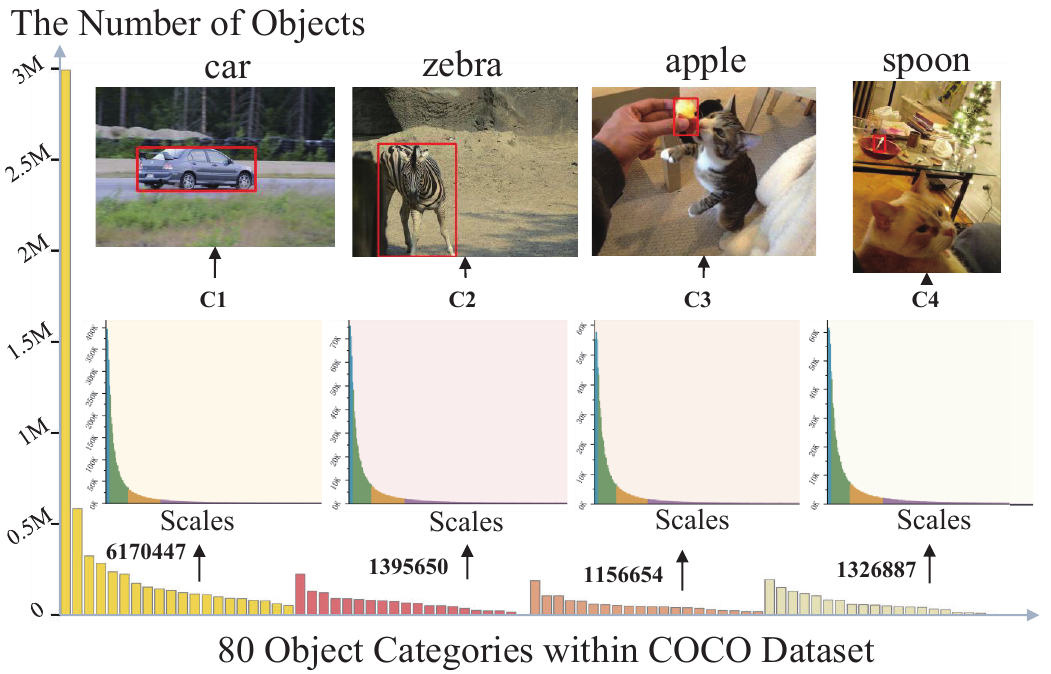}
    \end{center}
    \vspace{-2mm}
	\caption{Long-tailed distributions in object and scale in open-world scenarios: an analysis of four sets including 80 object classes. The x-axis represents scales and the y-axis represents the number of examples in the four sub-images.}
	\vspace{-4mm}
	\label{fig:motivate}
\end{figure}

In the context of general object retrieval, addressing the long-tailed distribution of objects is one of the primary challenges. This distribution pattern is a common occurrence in open-world VQA scenarios, as illustrated in~\cref{fig:motivate}. The majority of objects are concentrated within the head classes, making it more challenging to recognize or retrieve objects from the tail classes~\cite{liu2022open}. Furthermore, it's important to note that the scale distribution also exhibits a long-tailed pattern, as demonstrated in the sub-distributions of~\cref{fig:motivate}. This presents an additional challenge as it becomes difficult to learn effective embeddings for small scale objects under unsupervised conditions. 

In dynamic and open environments of VQA, most previously unseen objects belong to the tailed classes and possess small scales. This introduces challenges in the extraction and representation of such objects. While the segment anything model (SAM)~\cite{kirillov2023segment} offers a potential solution for capturing object spatial context, and unsupervised representation learning algorithms like DINO~\cite{caron2021emerging}, MoCo~\cite{chen2021}, iBOT~\cite{zhou2021ibot}, and STML~\cite{kim2022self} have demonstrated impressive performance in general image representation, employing these methods to obtain representations of open-vocabulary and small-scale objects may result in a degradation of performance. To tackle the long-tailed problem associated with objects and the corresponding scales in OR-OK-VQA, this paper introduces a multi-scale group collaborative embedding learning (MS-GCEL) method, designed to effectively learn object embeddings. The approach includes a potential object extraction network for object extraction and an MS-GCEL network for object embedding learning.

Furthermore, existing datasets used for deep feature embedding, such as ImageNet~\cite{russakovsky2015imagenet}, SOP~\cite{oh2016deep}, CUB~\cite{wah2011caltech}, and Cars~\cite{krause20133d}, primarily consist of images with uniform sizes. These samples may diverge considerably from those encountered in dynamic real-world VQA scenarios. Conversely, datasets employed for image retrieval tasks~\cite{hu2023co,ji2024hierarchical}, like Oxford5k~\cite{philbin2007object} and Paris6k~\cite{philbin2008lost}, enable the evaluation of retrieval performance for specific object categories. They fall short when it comes to training a model to represent a diverse range of objects in VQA. To better mimic real-world applications, we curate a new dataset by using the segment anything model (SAM)~\cite{kirillov2023segment} to extract objects from the amalgamating of COCO 2017~\cite{lin2014microsoft} and VOC 2007~\cite{everingham2007pascal} datasets. This dataset serves as a valuable resource for training and evaluating open-vocabulary object retrieval models.
Importantly, there are no strict category constraints imposed on either the training or test samples. Consequently, the model is trained on samples with uncertain categories and tested on samples with uncertain categories as well.
To assess the generalization capability of the model, we conducted additional object retrieval experiments across several datasets, including BelgaLogos~\cite{joly2009logo}, Visual Genome~\cite{krishna2017visual}, and LVIS~\cite{gupta2019lvis}. It is important to emphasize that these datasets were not used during the training phase. Furthermore, we curated the OK-VQA evaluation benchmark by incorporating images from the BelgaLogos, Visual Genome, and LVIS datasets to highlight the significance of object retrieval in VQA. The contributions of this work are summarized as follows:

\begin{itemize}
     \item We introduce the task of object retrieval for visual question answering with outside knowledge (OR-OK-VQA) and proposed a multi-scale group collaborative embedding learning (MS-GCEL) method for general object retrieval.

     \item We construct a benchmark of challenges by leveraging the SAM~\cite{kirillov2023segment} to extract objects from the amalgamating of COCO 2017~\cite{lin2014microsoft} and VOC 2007~\cite{everingham2007pascal} datasets, facilitating the training and evaluation of general object retrieval models. Furthermore, we assemble a rigorous test set specifically designed for evaluating open-set object retrieval performance. 

     \item In addition to our internally curated evaluation set, our object retrieval assessments encompass a variety of datasets which do not used during training, including BelgaLogos~\cite{joly2009logo}, Visual Genome~\cite{krishna2017visual}, and LVIS~\cite{gupta2019lvis}. Also, we curate an OR-OK-VQA evaluation benchmark to highlight the significance of object retrieval in VQA. %The results demonstrate that our proposed MS-GCEL method substantially enhances feature embedding performance when compared to popular methods, resulting in an object- and image-level mAP improvement of up to 6.69\% and 10.03\%, respectively.
     Evaluation on both general object retrieval and OR-OK-VQA tasks proved the effectiveness of the proposed method.
\end{itemize}
\section{Related Work}
\label{sec2}
The task of object retrieval for visual question answering with outside knowledge (OR-OK-VQA), which involves general object retrieval and OK-VQA, both remain a formidable challenge with a well-established research history. In the following subsections, we provide an overview of prior work related to this topic, encompassing object retrieval and OK-VQA.

\subsection{Object Retrieval}
\label{subsec2.1}
Early approaches frequently employed for object matching is to generate visual representations from the activations of convolutional layers~\cite{tolias2016particular} and transformer blocks~\cite{kan2022coded}. These studies extensively utilized deep metric learning methods, which aims to learn discriminative features from images by minimizing the distance between samples of the same class and maximizing the distance between samples of different classes. In early deep metric learning methods, contrastive loss~\cite{hadsell2006dimensionality} was effectively employed to optimize the pairwise distances between samples. To explore more intricate relationships between samples, various metric loss functions, including triplet loss~\cite{schroff2015facenet} and lifted structured loss~\cite{wang2019multi}, have been developed. These methods rely on supervised learning using image labels. However, in real-world object retrieval scenarios, objects are often unknown and lack labels. To learn representations of unlabeled objects, several techniques have been proposed. He \textit{et al.}~\cite{he2020momentum} introduced the momentum contrast (MoCo) method for unsupervised visual representation learning. Chen \textit{et al.}~\cite{chen2020simple} presented the simCLR framework, which is based on contrastive learning, for effective unsupervised visual representation learning.

In an effort to generate more robust pseudo-labels for unsupervised deep metric learning, Nguyen \textit{et al.}~\cite{nguyen2020deep} introduced a deep clustering loss to learn centroids. More recently, Kan \textit{et al.}~\cite{kan2021relative} proposed a relative order analysis (ROA) and optimization method for enhancing the relative ranking of examples in unsupervised deep metric learning. Li \textit{et al.}~\cite{li2021spatial} developed spatial assembly networks (SAN) to facilitate both supervised and unsupervised deep metric learning. Kim \textit{et al.}~\cite{kim2022self} introduced an effective teacher-student learning pipeline for embedding learning of unlabeled images.

The methods mentioned above have primarily been investigated using evenly distributed class examples with images of the same size. Their performance may deteriorate when applied to VQA. Our work aims to develop a general object embedding method that is more representative of the scenario of VQA, setting it apart from the aforementioned works.

\subsection{Visual Question Answering with Outside Knowledge}

With the rise of large language models (LLMs) and retrieval-augmented generation (RAG), VQA with external knowledge has become increasingly popular~\cite{zhao2023retrieving,zhao2024retrieval}. RAG addresses limitations of LLMs, particularly their tendency to produce `hallucinations' when handling queries outside their training data or requiring real-time information. Prior RAG-based approaches have focused on retrieving relevant text descriptions~\cite{lin2022retrieval} or identifying objects within large images to localize answer-relevant areas~\cite{wu2024v}. RAG methods are generally categorized into query-based, latent representation-based, logit-based, and speculative approaches~\cite{zhao2023retrieving}. In query-based RAG, user queries are enhanced with retrieved information, which is then fed into the generator's input—exemplified by RetrieveGAN~\cite{tseng2020retrievegan}, which integrates image patches and bounding boxes to improve relevance and precision. Latent representation-based RAG leverages cross-attention to fuse retrieved objects as latent features~\cite{chen2022re}. Logit-based RAG introduces retrieval data at the decoding stage~\cite{fei2021memory}, while speculative RAG optimizes response time by selectively combining retrieval with generation. Our object retrieval method for VQA falls under query-based RAG, where we focus on enhancing retrieval accuracy to improve VQA performance.

\section{Challenges in Object Retrieval for OK-VQA}
\label{sec:motivation}

Different from visual question answering (VQA), which answers a question based on the image content, VQA with outside knowledge (OK-VQA) requires retrieve external information that related with objects that can help to answer the question. Thus, one of the key tasks in OK-VQA is object retrieval. Unlike object detection~\cite{huang2024cross}, which aims to recognize~\cite{roshan2024neurocomputational} and locate objects~\cite{peng2024msednet} within an image, object retrieval is focused on ranking images and locating objects within the retrieved images. Typically, object detection algorithms are employed as the initial step in object retrieval to extract objects. In VQA scenarios, object retrieval introduces a new challenge: the query object can be any object, making the extraction and retrieval of arbitrary objects a challenging task. To address this challenge, previous state-of-the-art methods have introduced approaches like MViT~\cite{maaz2022class} and SAM~\cite{kirillov2023segment} for extracting both known and unknown objects. However, the representation of objects for retrieval remains a challenge that has not been thoroughly explored. To investigate this matter, we conducted experiments using the state-of-the-art unsupervised STML~\cite{kim2022self} method based on MViT and SAM. These experiments were conducted using our carefully curated training and validation object sets, which consist of a total of 80 classes from COCO 2017~\cite{lin2014microsoft} and VOC 2007~\cite{everingham2007pascal}.

\begin{figure*}[t]
    \begin{center}
    \includegraphics[width=\linewidth]{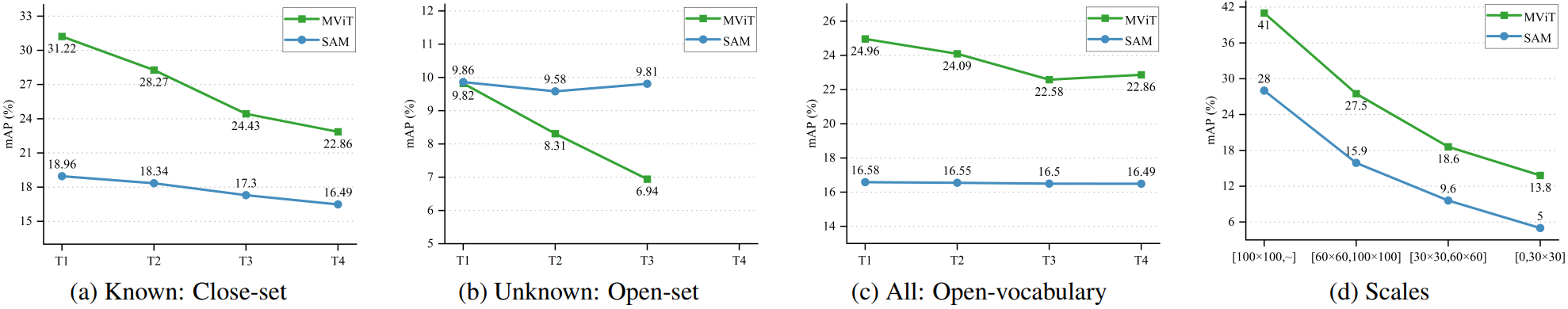}
    \end{center}
    % \vspace{-2mm}
	\caption{Retrieval scores (\%) at the object level with different object extraction methods on the validation set for object retrieval, considering known (close-set), unknown (open-set), all (open-vocabulary), and objects of different scales.}
	% \vspace{-4mm}
	\label{fig:challenges}
\end{figure*}

To simulate open-vocabulary object retrieval in OK-VQA scenario, we followed the dataset partitioning in~\cite{joseph2021towards} and divided the dataset into 4 tasks, denoted as Task1 to Task4, with each task progressively adding 20 new classes. The model is trained on each task and subsequently evaluated under three different scenarios: close-set (known), open-set (unknown), and open-vocabulary (all). In this context, ``known'' signifies that the test classes match the training classes, ``unknown'' indicates that the test classes are distinct from the training classes, and ``all'' denotes that the model is tested on all 80 classes. The results on the validation set are presented in~\cref{fig:challenges} (a)-(c). It is apparent that, as we progress from Task1 to Task4, the object-level retrieval mAP scores exhibit a consistent decrease for both the MViT and SAM extraction methods across all three scenarios. The observed decline in object-level retrieval mAP scores within the close-set and open-set scenarios can be attributed to the challenges posed by the introduction of new class samples. Additionally, for known objects (\cref{fig:challenges} (a)), the object-level mAP scores based on SAM are lower compared to those based on MViT. This can be attributed to the significantly higher number of objects extracted by SAM (averaging 89 per image) in comparison to MViT (averaging 14 per image).

\begin{figure*}[t]
    \begin{center}
    \includegraphics[width=\linewidth]{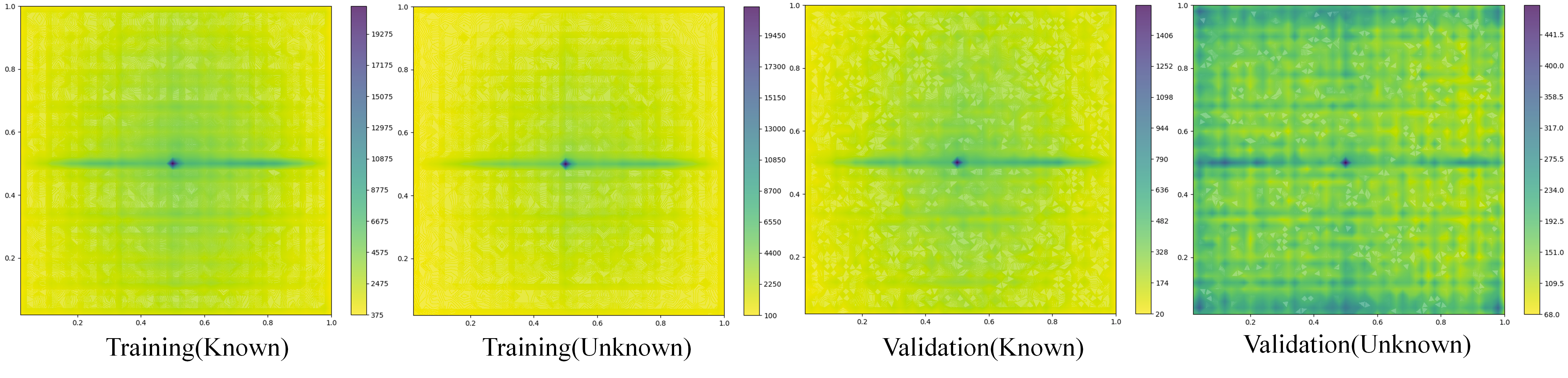}
    \end{center}
    \vspace{-2mm}
	\caption{Spatial distributions of known and unknown objects within both the training and validation sets for Task1.}
	\vspace{-4mm}
	\label{fig:motivation-distribution}
\end{figure*}

However, from~\cref{fig:challenges} (c), we have observed that the inclusion of more class samples also results in a decrease in object-level mAP scores within the open-vocabulary scenario. Upon conducting a thorough analysis of the underlying causes of this phenomenon, we have discerned two principal contributing factors. The first factor pertains to the inherent properties of the newly added classes, specifically their relatively smaller object sizes. This characteristic can exacerbate the challenges of unsupervised training by introducing a heightened susceptibility to labeling errors. As demonstrated in~\cref{fig:challenges} (d), the retrieval mAP score decreases with the smaller scale of the query object. The second factor pertains to the distinct spatial distribution characteristics exhibited by the newly incorporated classes in comparison to the known classes. As illustrated in~\cref{fig:motivation-distribution}, the spatial distributions of the unknown objects differ between the training and validation sets. These variations in spatial attributes further compound the challenges encountered within this OK-VQA scenario. In response to these challenges, we introduce a novel multi-scale group collaborative embedding learning method designed to learn models for general object representation. Furthermore, it is observed that the unknown object retrieval mAP scores (~\cref{fig:challenges} (b)) consistently outperform when utilizing the SAM for object extraction as compared to the MViT approach. Consequently, SAM is adopted as the preferred object extraction method in our experiments.

\section{Methods}
This paper focuses on achieving general object retrieval for OK-VQA. In the following subsections, we introduce our proposed method, multi-scale group collaborative embedding learning (MS-GCEL), to attain this objective.

% \begin{figure*}[htbp]
%     \begin{center}
%     \includegraphics[width=\linewidth]{figures/fig5_retrieval-pipeline-HR.pdf}
%     \end{center}
%     \vspace{-2mm}
% 	\caption{The pipeline for general object embedding learning using multi-scale group collaborative embedding learning (MS-GCEL) method.}
% 	\vspace{-4mm}
% 	\label{fig:framework}
% \end{figure*}

\begin{figure*}[htbp]
    \begin{center}
    \includegraphics[width=\linewidth]{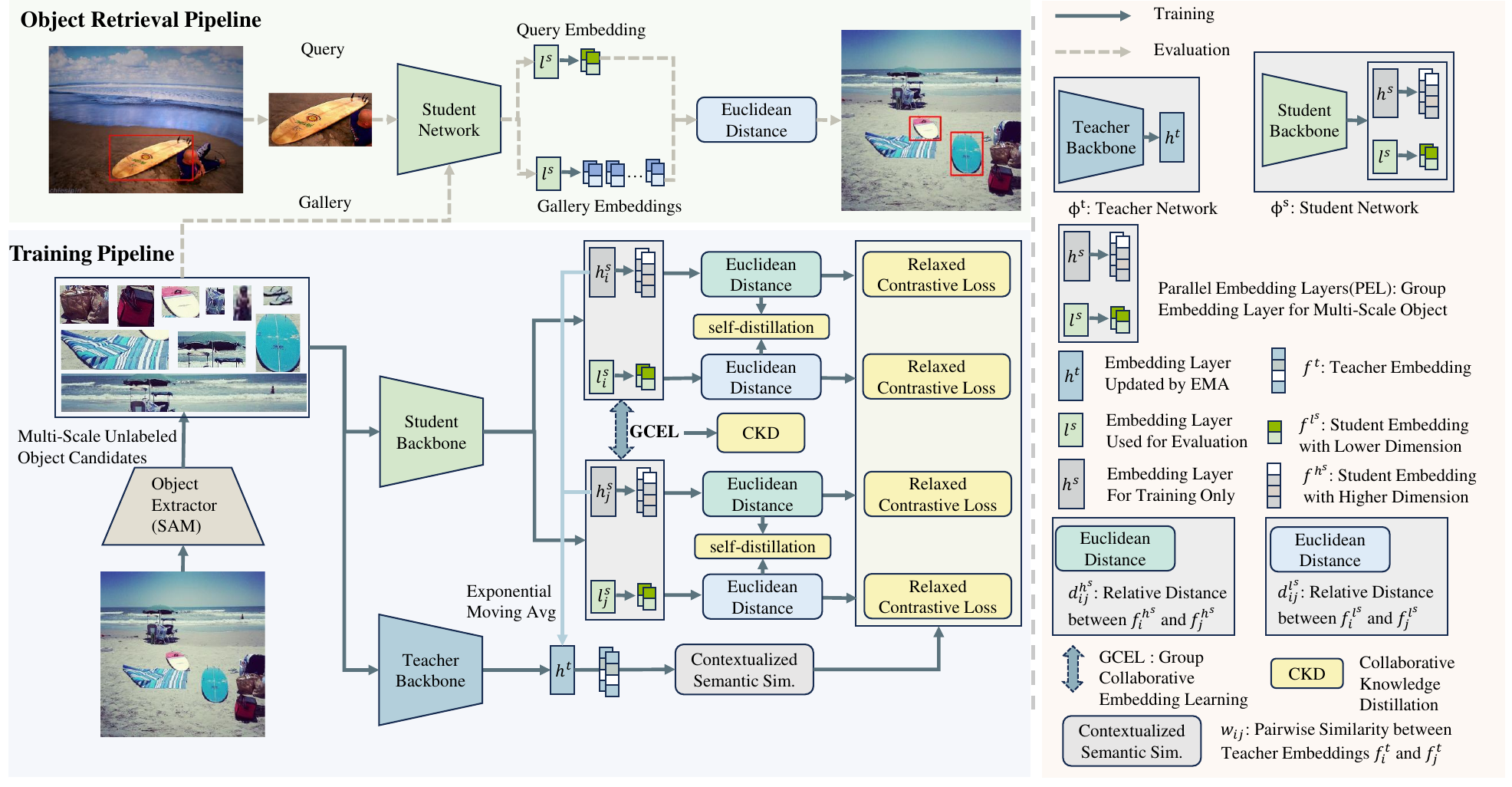}
    \end{center}
    \vspace{-4mm}
	\caption{An overview of our object retrieval and object embedding training pipeline. A teacher-student training strategy is employed during the training process. Firstly, we estimate contextualized semantic similarity between data pairs in the embedding space of the teacher network. Then, multi-scale group collaborative embedding learning is utilized to update the parameters of the student network. Ultimately, the trained student network is employed in the object retrieval process.}
	\vspace{-4mm}
	\label{fig:framework}
\end{figure*}

\subsection{Method Overview}

~\cref{fig:framework} offers an overview of the MS-GCEL method proposed for object retrieval. Our primary objective, as shown in the object retrieval pipeline, is to train a universal object embedding network capable of representing any object. This network enables the extraction of embeddings for both query and gallery objects. Subsequently, we compute the Euclidean distances between the query embedding and the gallery embeddings. These distances are used for ranking, resulting in the retrieval of the top-k objects, which are then located within their respective images. Finally, these retrieved images are used to assist VQA for LVLM. Our method includes a potential object extraction network for object extraction and an MS-GCEL network for object embedding learning. The potential object extraction network is the SAM~\cite{kirillov2023segment}. %Because SAM has been trained on a dataset of 10 million images, it exhibits the capability to perform hierarchical object segmentation, allowing for the extraction of various objects. Consequently, in this work, we harness the potential of SAM to extract objects from images, with the goal of achieving open-vocabulary object retrieval. 
The MS-GCEL network can be configured using popular CNN-based or Transformer-based networks. In our experiments, we explored various backbone networks. In the literature, the teacher-student training paradigm has been demonstrated as effective for embedding learning~\cite{kim2022self}. Consequently, we have also implemented the teacher-student training pipeline to train the MS-GCEL network. In this approach, the teacher network $\phi^{t}$ maintains the same architecture as the student network $\phi^{s}$, and its weights are updated with momentum using the exponentially moving average weights of the student network. The details of each part are described in the following sections.

\subsection{Multi-Scale Group Collaborative Embedding Learning}

\begin{figure}[tbp]
    \begin{center}
    \includegraphics[width=1.0\linewidth]{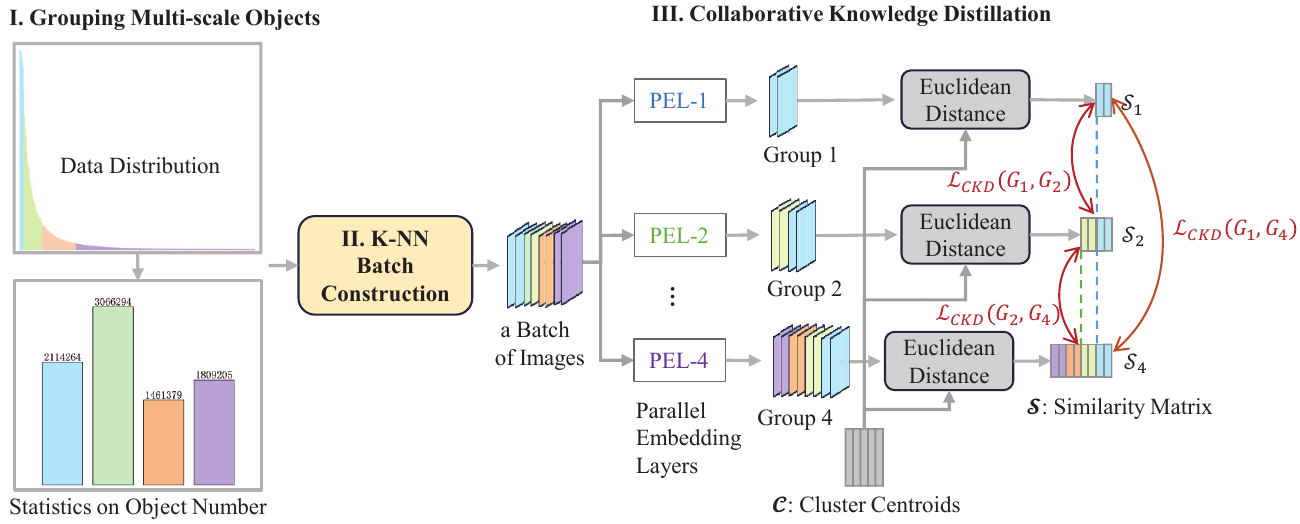}
    \end{center}
    \vspace{-4mm}
	\caption{The pipeline of multi-scale group collaborative embedding learning (MS-GCEL).}
	\vspace{-4mm}
	\label{fig:ms-gcel}
\end{figure}

~\cref{fig:ms-gcel} portrays the intricate pipeline of our MS-GCEL. As elaborated upon in~\cref{sec:motivation}, the effectiveness of embedding learning is closely associated with the accuracy of labeling, which, in turn, is linked to the scale of the image. Generally, images with larger scales tend to yield higher labeling accuracy, while those with smaller scales may result in lower labeling accuracy.

%To validate this observation, we conducted experiments involving model training with the exclusion of smaller-scale images from the training set. On the validation set, the object-level mAP scores are 12.8\%, 13.1\%, and 13.7\% when excluding images with sizes smaller than 10$\times$10, 20$\times$20, and 30$\times$30, respectively. However, when we exclude images with sizes larger than 30$\times$30, the mAP score decreased. It is evident that the best performance on the validation set is achieved by omitting images with scales smaller than 30$\times$30, suggesting that smaller image sizes may negatively impact the embedding model's performance.

Building upon the preceding analysis, we propose categorizing images according to their sizes. As depicted in the initial step of~\cref{fig:ms-gcel}, the data distribution exhibits a long-tailed nature. The horizontal axis represents the image sizes, arranged from small to large, while the vertical axis denotes the number of objects. To uniformly sample data across different scales, we segment this data into multiple groups with uniform distributions (e.g., 4), spanning from small to large based on image area.

Then, we explore the use of \textit{k}-nearest neighbors (K-NN) to sample data from each group, forming a batch of images, as illustrated in the second step of~\cref{fig:ms-gcel}. Subsequently, we compute the collaborative knowledge distillation loss based on the grouped embeddings. Specifically, for the $m$-th group, we denote the student output embedding vectors generated by $l^s_m$ as $\mathcal{F}^{l^s_m}$.  To initiate the process, we employ the k-means clustering algorithm to cluster all the training embeddings, resulting in a set of clustering centroids, denoted as $\mathcal{C}$. We then compute the normalized similarity matrix $\bm{S}_m$ between embeddings in group $m$ and cluster centroids in $\mathcal{C}$. For collaborative knowledge distillation (CKD) between group $a$ and group $b$, we utilize the Kullback-Leibler divergence to calculate their CKD loss, which is defined as follows
\begin{equation}
\begin{aligned}
    \mathcal{L}_{\text{CKD}}(G_a,G_b)=\frac{1}{|G_{a\cap b}|}\sum_{o_i\in G_{a\cap b}}\mathcal{L}_{\text{KL}}(\bm{S}^{i}_a,\bm{S}^{i}_b).\nonumber
\end{aligned}
\end{equation}
In the formula, $G_a$ and $G_b$ represent the object image sets in group $a$ and $b$, while $G_{a\cap b}$ denotes the set of identical object images found in both group $a$ and $b$. The term $|G_{a\cap b}|$ corresponds to the number of images in $G_{a\cap b}$.  $\bm{S}_a^{i}$ represents the indexed similarity vector between the embeddings of object $o_i$ and $\mathcal{C}$ within the matrix $\bm{S}_a$. The function of Kullback-Leibler divergence $\mathcal{L}_{\text{KL}}(\bm{S}^{i}_a,\bm{S}^{i}_b)$ is defined as follows
\begin{equation}
    \mathcal{L}_{\text{KL}}(\bm{S}^{i}_a,\bm{S}^{i}_b)=-\sum_{l=1}^L\bm{S}^{i}_a(l)\log\bm{S}^{i}_b(l),
\end{equation}
where $L=|\mathcal{C}|$, which is the total similarities between the embeddings of object $o_i$, and $\mathcal{C}$. Finally, the CKD loss in all groups is computed as follows
\begin{equation}
    \mathcal{L}_{\text{CKD}}=\frac{1}{\mathcal{|Q|}}\sum_{(a,b)\in \mathcal{Q}}\mathcal{L}_{\text{CKD}}(G_a,G_b),
\end{equation}
where $|Q|$ is calculated as $|Q|=\frac{{k(k-1)}}{2}$, with $k$ representing the number of groups.

\subsection{Overall Loss Function and OR-OK-VQA}
To train the network, we utilize a combination of knowledge distillation and contrastive loss functions. The primary objective of the knowledge distillation loss is to extract more informative features from the more dependable embeddings. Meanwhile, the contrastive loss function serves as a metric loss, with the aim of learning a Mahalanobis distance metric. They were defined according to the following definitions. First, we define the loss functions $\mathcal{L}^{\text{self}}_m$ and $\mathcal{L}^{\text{con}}_m$ following the works of STML~\cite{kim2022self}. Let the student output embedding vectors generated by $h^s_m$ and $l^s_m$ in the $m$-th group be denoted as $\mathcal{F}^{h^s_m}$ and $\mathcal{F}^{l^s_m}$. The self-distillation loss $\mathcal{L}^{\text{self}}_m$ is computed based on Kullback-Leibler divergence, as follows:
\begin{equation}
\mathcal{L}^{\text{self}}_m(\mathcal{F}^{h^s_m},\mathcal{F}^{l^s_m})=\sum_{i=1}^n\sum_{j\neq i}^n\frac{\psi(-d_{ij}^{l^s_m})}{n}\log\frac{\psi(-d_{ij}^{l^s_m})}{\psi(-d_{ij}^{h^s_m})},
\label{eq:self}
\end{equation}
where $d_{ij}^{l^s_m}:=||\bm{f}_{i}^{l^s_m}-\bm{f}_{j}^{l^s_m}||_2$ / $(\frac{1}{n}\sum_{k=1}^n||\bm{f}_{i}^{l^s_m}-\bm{f}_{k}^{l^s_m}||_2$) represents the relative distance between $\bm{f}_{i}^{l^s_m}$ and $\bm{f}_{j}^{l^s_m}$. Similarly, $d_{ij}^{h^s_m}$ represents the relative distance between $\bm{f}_{i}^{h^s_m}$ and $\bm{f}_{j}^{h^s_m}$. The function $\psi(\cdot)$ corresponds to the softmax operation, and $n$ represents the number of samples in the $m$-th group. The relaxed contrastive loss $\mathcal{L}^{\text{con}}_m$ is defined as follows
\begin{equation}
\begin{aligned}
    \mathcal{L}^{\text{con}}_m(\mathcal{F}^{h^s_m})=&\frac{1}{n}\sum_{i=1}^{n}\sum_{j\neq i}^nw_{ij}(d_{ij}^{h^s_m})^2&\\+&\frac{1}{n}\sum_{i=1}^n\sum_{j \neq i}^n(1-w_{ij})[\delta-d_{ij}^{h^s_m}]_{+}^2,
\end{aligned}
\label{eq:contrastive}
\end{equation}
where $w_{ij}=exp(-\frac{||\bm{f}_{i}^{t}-\bm{f}_{j}^{t}||_2^2}{\sigma})$ represents a pairwise similarity score between teacher embeddings $\bm{f}_{i}^{t}$ and $\bm{f}_{j}^{t}$, and $\delta$ is a margin. It's worth noting that the same training approach is applied to $l^s_m$ as well.
The overall loss function $\mathcal{L}_{\text{MS-GCEL}}$ is defined as follows
\begin{equation}
\begin{aligned}
    \mathcal{L}_{\text{MS-GCEL}}=&\sum_{i=1}^k\mathcal{L}^{\text{self}}_i(\mathcal{F}^{h^s_i},\mathcal{F}^{l^s_i})+\sum_{i=1}^k\mathcal{L}^{\text{con}}_i(\mathcal{F}^{h^s_i})& \nonumber\\
    +&\sum_{i=1}^k\mathcal{L}^{\text{con}}_i(\mathcal{F}^{l^s_i})+\mathcal{L}_{\text{CKD}}.
\end{aligned}
\end{equation}

After training the network, the OR-OK-VQA task is performed in a human–computer interaction manner. First, gallery image features are extracted and indexed using the learned network with the MS-GCEL loss function. Given an input image and question, we crop a question-relevant object image and extract its features using the trained network. These object features are then used to retrieve relevant images from the gallery. Finally, the retrieved images and the question are fed into the large vision-language model (LVLM) to assist with the VQA task.
\section{Experiments}
\subsection{Datasets and Implementation Details}

Our training dataset is curated from pre-existing open-access object detection datasets, i.e.,  COCO 2017 \cite{lin2014microsoft} and VOC 2007 \cite{everingham2007pascal}. The testing dataset consists of publicly available object retrieval and object detection datasets, which serve to evaluate the performance of general object retrieval. We also established an OK-VQA evaluation benchmark using images from the BelgaLogos~\cite{joly2009logo}, Visual Genome~\cite{krishna2017visual}, and LVIS~\cite{gupta2019lvis} datasets.  The details of our experimental datasets are summarized in ~\cref{tab:dataset_statistics}. In the following, we provide a simple description of the datasets.
%Their details are summarised in Table \ref{tab:dataset_statistics} and detailed information on these datasets is provided below. 

\begin{table}[t]
		\caption{Statistics on the number of training and test sets.}
 		\vspace{-6mm}
            \renewcommand\arraystretch{1.0}
		\begin{center}
		\resizebox{0.8\linewidth}{!}{
			\begin{tabularx}{7.5cm}{lcc}
				\hline 
				 & Images& Objects\\
				\hline
                  training dataset & 121,298 & 8,035,395\\
                  COCO-VOC test query & 9,904 & 51,757\\
                  COCO-VOC test gallery &9,904 &862,861\\
                  BelgaLogos test query&55&55\\
                  BelgaLogos test gallery&10,000&939,361\\
                  LVIS test query &4726 &50,537\\
                  LVIS test gallery &4726 &433,671\\
                  Visual Genome test query &108,077 &79,921\\
                  Visual Genome test gallery &108,077 &9,803,549\\
				\hline
			\end{tabularx}
			}
		\end{center}
		\vspace{-7mm}
		\label{tab:dataset_statistics}
	\end{table}

\textbf{General object retrieval dataset.} The training and validation sets are constructed by extracting objects from the training sets of COCO 2017 and VOC 2007, as well as the validation set of VOC 2007. The training set comprises 121,298 images, while the validation set contains 2,000 images. The test set is composed of objects extracted from the validation set of COCO 2017 and the test set of VOC 2007, totaling 9,904 images. To perform object extraction, we employed SAM \cite{kirillov2023segment}. The resulting partitioned sets consist of more than 8 million objects for the training set, 173K for the validation set, and 862K for the test set. In the zero-shot object retrieval setup, we partition the training set into four distinct groups, with each group comprising 20 object classes. These four group datasets are employed to create four tasks, with each task progressively introducing an additional set of 20 new classes. The classes utilized for training (known classes highlighted in light yellow) and testing (unknown classes highlighted in light blue) in each task are presented in Table \ref{tab:tasksplit}. In Task 1, the training set consists of the most popular 20 classes, while the remaining 60 classes are used for testing. For Task 2, 20 new classes are introduced along with the original Task 1 classes for training, and the remaining 40 classes are designated for testing. Task 3 involves the addition of another set of 20 new classes to Task 2, with the remaining 20 classes used for testing. Finally, Task 4 incorporates all 80 classes for both training and testing phases. 

% The model is trained using one or more of these groups and subsequently evaluated on the remaining groups, as shown in \cref{fig:challenges} (b).

\textbf{Generalization evaluation dataset.} To assess the generalization ability of the trained MS-GCEL model, we conducted evaluations on the BelgaLogos \cite{joly2009logo}, Visual Genome \cite{krishna2017visual}, and LVIS \cite{gupta2019lvis} datasets. The BelgaLogos dataset consists of 10K images and 55 query logos, primarily used to evaluate the logo retrieval performance of an algorithm. The provided 55 query logo images of the BelgaLogos dataset are utilized as the query set, while the gallery set comprises 939,361 objects extracted using SAM\cite{kirillov2023segment}. In the case of the Visual Genome dataset, which contains 108K images and 2,516,939 labeled bounding boxes, we define a set of 79,921 objects as the query set and directly performed object retrieval using the trained model on the whole dataset. As for the LVIS dataset, it shares the same images as the COCO 2017 dataset but features 1000 class labels. We compute evaluation scores based on the labeled objects in the validation set of the LVIS dataset.

\textbf{OK-VQA evaluation dataset.} To evaluate the performance of OK-VQA, we constructed a specialized VQA dataset by selecting images from the BelgaLogos, Visual Genome, and LVIS datasets. A total of 36 questions are crafted for these images, with most answers requiring external knowledge beyond the information directly available in the images. To correctly answer these questions, relevant images must be retrieved from external datasets to provide additional contextual information.

\textbf{Evaluation Metrics.} We use Recall@1 and mean average precision (mAP) as the evaluation metrics for object retrieval. The Recall@1 is defined as follows: Recall@1 = $\frac{1}{q}\sum_{i=1}^{q}r_i$, where $q$ represents the number of queries, and $r_i$ is equal to 1 if the top-1 returned objects contain the ground truth object, and 0 otherwise.
The mAP is calculated as: mAP = $\frac{1}{q}\sum_{i=1}^{q}\sum_{j=1}\frac{j}{p_j}$, where $q$ represents the number of queries, and $p_j$ is the ranking index of a returned object that matches the ground truth. We calculate image-level and object-level retrieval scores. The image-level score is computed using an IoU threshold of $1e^{-10}$, and the object-level score is computed using an IoU threshold of 0.3.

\textbf{Implementation Details.}
%In our experiments utilizing SAM~\cite{kirillov2023segment} and MViT~\cite{maaz2022class}, different settings for object extraction were employed. Specifically, the ViT-H backbone network was utilized for object extraction based on SAM. For object extraction based on MViT, the text prompts 'all objects,' 'all entities,' 'all visible entities and objects,' and 'all obscure entities and objects' were employed. More training details are provided in ~\cref{sec:imple_detail}.
In our experiments, we utilize SAM \cite{kirillov2023segment} and MViT \cite{maaz2022class} for object extraction. Specifically, the ViT-H backbone network was utilized for object extraction based on SAM. For object extraction based on MViT, the text prompts 'all objects,' 'all entities,' 'all visible entities and objects,' and 'all obscure entities and objects' were employed. 
Our teacher-student network is trained on two RTX 3090 (24GB) GPUs with a batch size of 120. The nearest neighbors are set to 5. Each batch contains images of various sizes for MS-GCEL learning. The group number of the MS-GCEL is set to 4, and the clustering number is set to 100 unless specified otherwise. The dimension of teacher embeddings $f^t$ was set to 1024, while the dimensions of student embeddings $f^{l^s}$ and $f^{h^s}$ were both set to 512. The $\sigma$ to calculate $w_{ij}$ and the margin $\delta$ are 3 and 1 respectively. The initial learning rate is set as $10^{-4}$ on the backbone of GoogLeNet, and $3 \times 10^{-5}$ for MiT-B2 and ViT-B/16, both of which are scaled down by the cosine decay function\cite{loshchilov2016sgdr}. During training, The Adamp\cite{heo2020adamp} optimizer was employed with a weight decay set to 0.01. and the K-nearest neighbor sampling distance matrix was updated once every 1000 iterations.

%Our teacher-student network is trained on two RTX 3090 (24GB) GPUs with a batch size of 120. The nearest neighbors are set to 5. Each batch contains images of various sizes for MS-GCEL learning. The group number of the MS-GCEL is set to 4, and the clustering number is set to 100 unless specified otherwise. The dimension of teacher embeddings $f^t$ was set to 1024, while the dimensions of student embeddings $f^{l^s}$ and $f^{h^s}$ were both set to 512. The $\sigma$ to calculate $w_{ij}$ and the margin $\delta$ in Equation \ref{eq:contrastive} are 3 and 1 respectively. The initial learning rate is set as $10^{-4}$ on the backbone of GoogLeNet, and $3 \times 10^{-5}$ for MiT-B2 and ViT-B/16, both of which are scaled down by the cosine decay function~\cite{loshchilov2016sgdr}. During training, The Adamp~\cite{heo2020adamp} optimizer was employed with a weight decay set to 0.01. and the K-nearest neighbor sampling distance matrix was updated once every 1000 iterations.

%  \begin{figure*}[t]
%     \begin{center}
%     \includegraphics[width=0.9\linewidth]{figures/fig7_supplemental_compare_examples.pdf}
%     \end{center}
%     \vspace{-4mm}
% 	\caption{The comparison of three object retrieval examples between the MS-GCEL and STML~\cite{kim2022self} methods. Retrieved objects with \textcolor{green}{green} boxes are correct ones with the same class label as the query object. Those with \textcolor{red}{red} boxes are incorrect results.}
% 	\vspace{-8mm}
% 	\label{fig:compare_case.png}
% \end{figure*}

\subsection{Effectiveness Verification of the MS-GCEL Method}
To assess the effectiveness of the MS-GCEL method, we conducted experiments using different backbone networks and various initialization methods. The results on the curated test set are presented in~\cref{tab:effectiveness}. In this table, methods based on the GoogLeNet and MiT-B2 backbone networks are initialized with parameters trained on the ImageNet-1K~\cite{russakovsky2015imagenet} dataset. Our method based on the ViT-B/16~\cite{dosovitskiy2020image} backbone network are initialized using parameters trained by the MoCo-V3~\cite{chen2021} and CLIP~\cite{radford2021learning} methods, respectively, and then fine-tuned on our curated training dataset using the proposed MS-GCEL method.

From~\cref{tab:effectiveness}, it is evident that our method outperforms the state-of-the-art unsupervised STML~\cite{kim2022self} method, achieving an improvement of 2.65\% and 10.03\% in image level mAP for GoogLeNet and MiT-B2 backbone networks, respectively. %Additionally, it is notable that the general object retrieval score exhibits a significant disparity compared to the other metrics, underscoring the substantial potential for enhancement within the domain of object retrieval. The MS-GCEL method with the MiT-B2 backbone network obtained the highest object level mAP score.
To further verify the effect of multi-scale group collaborative embedding learning (MS-GCEL), we present the results for different sizes of query objects, as shown in~\cref{tab:coco_MS_results_google} and~\cref{tab:coco_MS_results_MiT}. Specifically,
~\cref{tab:coco_MS_results_google} reports object retrieval results based on the GoogLeNet backbone, while~\cref{tab:coco_MS_results_MiT} presents results using the MiT backbone. We can see that the MS-GCEL obtained better performance in both the object- and image-level retrieval scores for different sizes of query objects. Moreover, the larger size of objects, the higher the performance for both the STML~\cite{kim2022self} and the MS-GCEL methods.

\begin{table}[th]
		\caption{Retrieval scores (\%) at both the object level (O-) and image level (I-) with different backbone networks and different initialization methods on the object retrieval test set.}
 		\vspace{-6mm}
            \renewcommand\arraystretch{1.1}
		\begin{center}
		\resizebox{1.0\linewidth}{!}{
			\begin{tabularx}{10.8cm}{ll|cccc}
				\hline 
				 Backbone&Methods& O-R@1&O-mAP&I-R@1& I-mAP\\
				\hline
                  \multirow{2}{*}{GoogLeNet~\cite{szegedy2015going}}&STML~\cite{kim2022self} & 65.93 &14.89&69.70&65.04\\
                  &\textbf{MS-GCEL}& \textbf{68.17}& \textbf{15.37} &\textbf{71.90}& \textbf{67.69}\\
                  \multirow{2}{*}{MiT-B2~\cite{xie2021segformer}}&STML~\cite{kim2022self} &58.62  &17.48 &66.86 &58.60 \\
                  &\textbf{MS-GCEL}& \textbf{68.60}& \textbf{17.64} &\textbf{73.40}& \textbf{68.63}\\
                  \hline
                   \multirow{4}{*}{ViT-B/16~\cite{dosovitskiy2020image}}&MoCo-V3~\cite{chen2021} & 77.40 &15.94&81.29&81.26\\
                  &\textbf{MS-GCEL}& \textbf{77.88}& \textbf{16.84} &\textbf{82.04}& \textbf{82.12}\\
                  &CLIP~\cite{radford2021learning}& 68.44 &14.56&74.01&70.30\\
                  &\textbf{MS-GCEL}& \textbf{69.07}& \textbf{16.07} &\textbf{74.49}& \textbf{70.97}\\
				\hline
			\end{tabularx}
			}
		\end{center}
		\vspace{-6mm}
		\label{tab:effectiveness}
	\end{table}

 \begin{table*}[th]
		\caption{Retrieval scores (\%) of different scales of query objects based on the GoogLeNet backbone.}
 		\vspace{-6mm}
            \renewcommand\arraystretch{1.05}
		\begin{center}
		\resizebox{0.9\linewidth}{!}{
			\begin{tabularx}{17.1cm}{cl|cccc|cccc}
				\hline 
				 % Sizes&Methods& O-R@1&O-mAP&I-R@1& I-mAP\\
                 \multirow{2}{*}{Sizes}&\multirow{2}{*}{Methods}&\multicolumn{4}{c|}{COCO-VOC}&\multicolumn{4}{c}{LVIS}\\
                 \cline{3-10}
                 && O-R@1&O-mAP&I-R@1& I-mAP& O-R@1&O-mAP&I-R@1& I-mAP\\
				\hline
                  \multirow{2}{*}{[0,$20\times 20$]}&STML~\cite{kim2022self} & 14.30 &1.63&20.67&15.23&21.25&2.80&22.10&24.66\\
                  &\textbf{MS-GCEL}& \textbf{18.38}& \textbf{1.68} &\textbf{24.06}& \textbf{20.27}& \textbf{25.18}& \textbf{3.29} &\textbf{26.43}& \textbf{28.98}\\
                  \multirow{2}{*}{[$20 \times 20$,$30 \times 30$]}&STML~\cite{kim2022self} &42.72  &3.61 &49.25 &44.50&52.48&5.88&53.22&57.72 \\
                  &\textbf{MS-GCEL}& \textbf{52.20}& \textbf{3.75} &\textbf{57.11}& \textbf{53.64}&\textbf{58.77}&\textbf{7.01}&\textbf{59.58}&\textbf{63.89}\\
                   \multirow{2}{*}{[$30 \times 30$,$60 \times 60$]}%&DINO~\cite{caron2021emerging} & 77.25 &17.33&81.08&81.38\\
                  %&\textbf{MS-GCEL}& \textbf{}& \textbf{} &\textbf{}& \textbf{}\\
                  &STML~\cite{kim2022self} & 64.51 &6.91&69.22&64.83&67.22&7.40&67.77&72.46\\
                  &\textbf{MS-GCEL}& \textbf{68.13}& \textbf{6.96} &\textbf{72.89}& \textbf{68.50}&\textbf{70.46}&\textbf{8.25}&\textbf{71.07}&\textbf{75.72}\\
                  \multirow{2}{*}{[$60 \times 60$,$100 \times 100$]}
                  &STML~\cite{kim2022self}& 78.05 &14.73&81.83&76.06&76.14&8.46&76.85&81.25\\
                  &\textbf{MS-GCEL}& \textbf{79.74}& \textbf{15.34} &\textbf{83.32}& \textbf{78.25}&\textbf{78.47}&\textbf{9.24}&\textbf{79.32}&\textbf{83.53}\\
                   \multirow{2}{*}{[$100 \times 100$,$\sim$]}
                  &STML~\cite{kim2022self}& 85.74 &26.68&87.66&83.50&81.87&11.30&82.77&86.54\\
                  &\textbf{MS-GCEL}& \textbf{87.56}& \textbf{27.77} &\textbf{89.61}& \textbf{86.47}&\textbf{83.71}&\textbf{13.01}&\textbf{85.00}&\textbf{88.74}\\
				\hline
			\end{tabularx}
			}
		\end{center}
		\vspace{-6mm}
		\label{tab:coco_MS_results_google}
	\end{table*}
\begin{table*}[t]
		\caption{Retrieval scores (\%) of different scales of query objects based on the MiT backbone.}
 		\vspace{-6mm}
            \renewcommand\arraystretch{1.05}
		\begin{center}
		\resizebox{0.9\linewidth}{!}{
			\begin{tabularx}{17.1cm}{cl|cccc|cccc}
				\hline 
                  \multirow{2}{*}{Sizes}&\multirow{2}{*}{Methods}&\multicolumn{4}{c|}{COCO-VOC}&\multicolumn{4}{c}{LVIS}\\
                 \cline{3-10}
                 && O-R@1&O-mAP&I-R@1& I-mAP& O-R@1&O-mAP&I-R@1& I-mAP\\
				\hline
                  \multirow{2}{*}{[0,$20\times 20$]}&STML~\cite{kim2022self} & 11.85 &1.39&18,67&12.75&16.18&2.36&17.49&20.68\\
                  &\textbf{MS-GCEL}& \textbf{15.19}& \textbf{1.71} &\textbf{22.24}& \textbf{16.45}&\textbf{21.67}&\textbf{3.01}&\textbf{22.80}&\textbf{25.78}\\
                  \multirow{2}{*}{[$20 \times 20$,$30 \times 30$]}&STML~\cite{kim2022self} &34.82  &\textbf{4.55} &43.88 &35.67&42.99&5.56&44.59&49.30 \\
                  &\textbf{MS-GCEL}& \textbf{46.11}& 3.72 &\textbf{52.56}& \textbf{47.93}&\textbf{57.16}&\textbf{7.03}&\textbf{58.39}&\textbf{62.55}\\
                   \multirow{2}{*}{[$30 \times 30$,$60 \times 60$]}
                  &STML~\cite{kim2022self} & 54.33 &11.07&63.39&52.57&61.55&7.93&63.27&68.25\\
                  &\textbf{MS-GCEL}& \textbf{68.33}& \textbf{8.78} &\textbf{73.97}& \textbf{67.92}&\textbf{71.35}&\textbf{8.83}&\textbf{72.20}&\textbf{76.56}\\
                  \multirow{2}{*}{[$60 \times 60$,$100 \times 100$]}
                  &STML~\cite{kim2022self}& 67.86 &19.39&77.06&65.58&66.30&10.41&70.39&74.52\\
                  &\textbf{MS-GCEL}& \textbf{80.86}& \textbf{18.25} &\textbf{85.61}& \textbf{79.11}&\textbf{79.04}&\textbf{11.03}&\textbf{80.31}&\textbf{84.15}\\
                   \multirow{2}{*}{[$100 \times 100$,$\sim$]}
                  &STML~\cite{kim2022self}& 78.72 &29.15&86.41&80.10&72.51&15.54&77.61&82.67\\
                  &\textbf{MS-GCEL}& \textbf{87.86}& \textbf{31.39} &\textbf{91.02}& \textbf{87.98}&\textbf{83.93}&\textbf{16.65}&\textbf{86.17}&\textbf{90.16}\\
				\hline
			\end{tabularx}
			}
		\end{center}
		\vspace{-6mm}
		\label{tab:coco_MS_results_MiT}
	\end{table*}

\begin{table*}[t]
		\caption{Retrieval scores (\%) at both the object level (O-) and image level (I-) on different datasets.}
 		\vspace{-6mm}
            \renewcommand\arraystretch{1.0}
		\begin{center}
		\resizebox{\linewidth}{!}{
			\begin{tabularx}{21.4cm}{ll|c|cccc|cccc}
              \hline
				 \multirow{2}{*}{Backbone}&\multirow{2}{*}{Methods}&\multicolumn{1}{c|}{BelgaLogos}&\multicolumn{4}{c|}{Visual Genome}&\multicolumn{4}{c}{LVIS}\\
               \cline{3-11}
                   && I-mAP&O-Recall@1& O-mAP&I-Recall@1& I-mAP&O-Recall@1& O-mAP&I-Recall@1& I-mAP\\
				\hline
                  \multirow{2}{*}{GoogLeNet~\cite{szegedy2015going}}&STML~\cite{kim2022self} &63.61 &31.82&28.56&31.19&43.54 &55.79&6.72&56.57&60.42\\
                  &\textbf{MS-GCEL}& \textbf{65.93}& \textbf{34.01} &\textbf{30.39}& \textbf{34.11} & \textbf{46.93} &\textbf{58.75}& \textbf{7.55} & \textbf{59.75} &\textbf{63.47}  \\
                  \multirow{2}{*}{MiT-B2~\cite{xie2021segformer}}&STML~\cite{kim2022self} &62.71&26.88&25.26&28.34 &43.10  &49.31&7.48&51.65&56.14 \\
                  &\textbf{MS-GCEL}& \textbf{64.59} & \textbf{33.39}  & \textbf{31.95} &\textbf{33.59}& \textbf{46.80}  & \textbf{56.83} & \textbf{8.37} & \textbf{58.11} & \textbf{61.96}\\
                  \hline
                   \multirow{4}{*}{ViT-B/16~\cite{dosovitskiy2020image}}%&DINO~\cite{caron2021emerging} & 82.30 &42.52&39.12&43.58&66.69 &64.65&11.10&66.51&71.12\\
                  %&\textbf{MS-GCEL}& & \textbf{} &\textbf{}& \textbf{}\\
                  &MoCo-V3~\cite{chen2021} & 86.32 &\textbf{43.13}&\textbf{39.47}&\textbf{44.07}&\textbf{67.33} &64.95&11.04&\textbf{67.12}&71.51\\
                  &\textbf{MS-GCEL}& \textbf{86.69}& 42.77 &38.95& 43.67 & 66.67 &\textbf{65.14}& \textbf{11.19} & 67.10 &\textbf{71.59}\\
                  &CLIP~\cite{radford2021learning}& 82.01 &33.15&30.11&33.62&51.28 &55.28&8.51&57.07&61.45 \\
                  &\textbf{MS-GCEL}& \textbf{83.73}& \textbf{33.41} &\textbf{30.35}& \textbf{33.75} &\textbf{51.80} &\textbf{55.69} &\textbf{8.70}& \textbf{57.44} &\textbf{61.95}\\
				\hline
			\end{tabularx}
			}
		\end{center}
		\vspace{-6mm}
		\label{tab:comparison}
	\end{table*}

%~\cref{fig:compare_case.png} illustrates the comparison of three object retrieval examples between the MS-GCEL and STML~\cite{kim2022self} methods. The first retrieval example is based on the GoogLeNet backbone network, while the second and third retrieval examples are based on the MiT-B2 backbone network. These examples demonstrate that the MS-GCEL method is capable of returning more relevant objects associated with the query object.

\subsection{Comparison and Analysis of General Object Retrieval and OK-VQA}
Moreover, we conduct evaluation experiments on the BelgaLogos~\cite{joly2009logo}, LVIS~\cite{gupta2019lvis} and Visual Genom datasets using the trained model. Results are shown in~\cref{tab:comparison}. We have observed that our approach using GoogLeNet~\cite{szegedy2015going} and MiT-B2~\cite{xie2021segformer} backbone networks effectively enhances retrieval performance across various datasets. The object-level mAP scores demonstrated a 6.69\% improvement on the Visual Genome dataset and a 0.89\% improvement on the LVIS dataset when using the MiT-B2 backbone network. The highest mAP scores and Recall@1 scores on the BelgaLogos, Visual Genome, and LVIS dataset are achieved with the parameters initialized by the MoCo-V3. However, on the Visual Genome dataset, the performance of the MS-GCEL method with MoCo-V3 slightly decreases due to the lower-dimensional feature embeddings (512 v.s. 1000).

\begin{figure}[t]
    \begin{center}
    \includegraphics[width=1.0\linewidth]{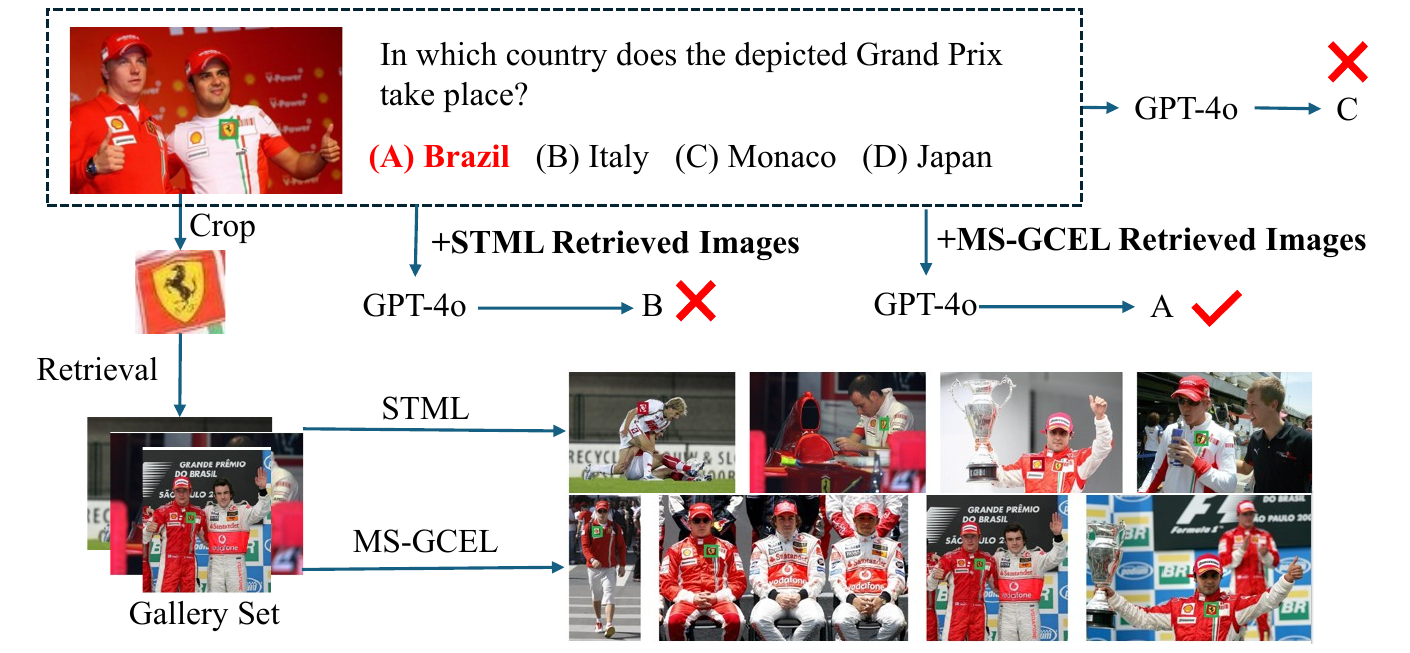}
    \end{center}
    \vspace{-4mm}
	\caption{VQA examples using GPT-4o, comparing results without and with the integration of STML and MS-GCEL methods.}
	\vspace{-4mm}
	\label{fig:Results-case}
\end{figure}

\begin{figure*}[!t]
    \begin{center}
    \includegraphics[width=0.8\linewidth]{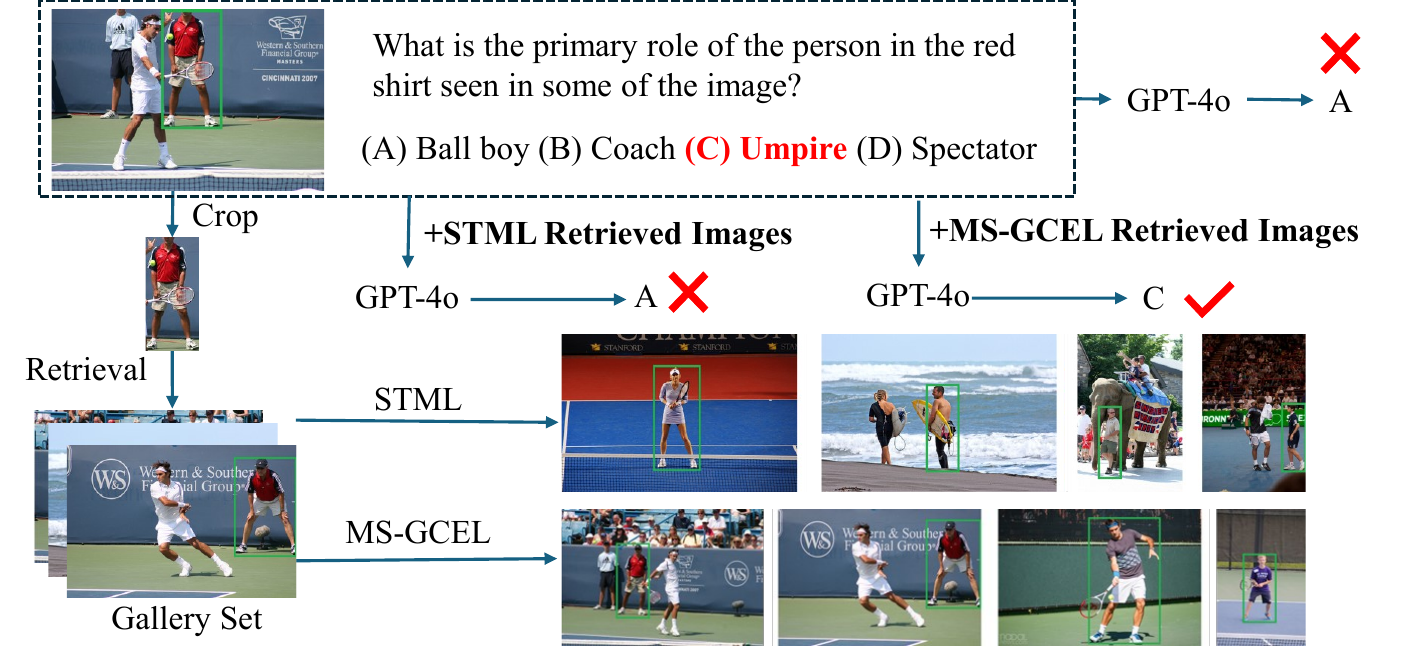}
    \end{center}
    \vspace{-4mm}
	\caption{VQA examples using GPT-4o, comparing results without and with the integration of STML and MS-GCEL methods.}
	% \vspace{-8mm}
	\label{fig:ok-vqa-1}
\end{figure*}

\begin{figure*}[t]
    \begin{center}
    \includegraphics[width=0.8\linewidth]{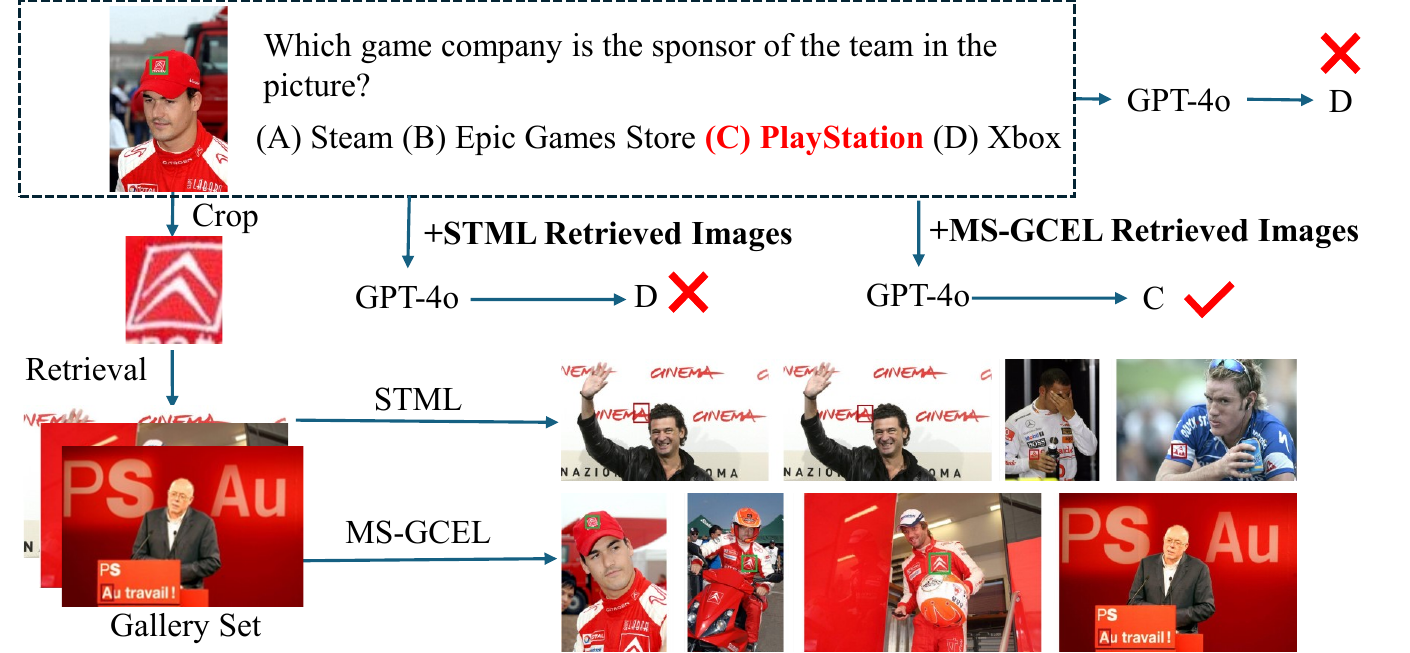}
    \end{center}
    \vspace{-2mm}
	\caption{VQA examples using GPT-4o, comparing results without and with the integration of STML and MS-GCEL methods.}
	\vspace{-4mm}
	\label{fig:ok-vqa-2}
\end{figure*}

The object retrieval results for OK-VQA on the curated evaluation dataset are shown in ~\cref{tab:or-ok-vqa}. We observe that enhancing the object retrieval model leads to improved VQA performance.~\cref{fig:Results-case},~\cref{fig:ok-vqa-1} and~\cref{fig:ok-vqa-2} present VQA examples using GPT-4o, comparing results without and with the integration of STML and MS-GCEL methods. We can see that incorporating MS-GCEL results enables GPT-4o to answer the question accurately.

%~\cref{fig:retrieval-examples} provides logo retrieval example showcasing the effectiveness of the proposed MS-GCEL method. These examples illustrate the ability to ground multiple objects within the retrieved images. 

% \begin{figure*}[t]
%     \begin{center}
%     \includegraphics[width=0.9\linewidth]{figures/fig8_retrieval-examples-multi-HR.png}
%     \end{center}
%     \vspace{-2mm}
% 	\caption{Logo retrieval example on the BelgaLogos using the proposed MS-GCEL method.}
% 	\vspace{-8mm}
% 	\label{fig:retrieval-examples}
% \end{figure*}

\begin{table}[t]
		\caption{OR-OK-VQA scores (\%) with the GPT-4o  based on the STML and MS-GCEL object retrieval methods.}
 		\vspace{-6mm}
            \renewcommand\arraystretch{1.05}
		\begin{center}
		\resizebox{1.0\linewidth}{!}{
			\begin{tabularx}{8.0cm}{l|ccc}
				\hline 
				 Methods&Without Retrieval&STML&MS-GCEL\\
				\hline
                  Scores & 25.00 & 33.33 &\textbf{36.11}\\
				\hline
			\end{tabularx}
			}
		\end{center}
		\vspace{-6mm}
		\label{tab:or-ok-vqa}
	\end{table}

\subsection{Ablation Study}
We conducted ablation studies on our curated object retrieval dataset using the GoogLeNet~\cite{szegedy2015going} backbone. In the subsequent ablation experiments, we utilize objects extracted by SAM for both training and validation. 

% (1) \textbf{Impact of the clustering number.} We examined the impact of the clustering number on deep feature embedding performance, with the number of groups fixed at 3. The results on the validation set are presented in~\cref{tab:ab-cluster}, and it's evident that the best performance is achieved when the number of clusters is set to 100.

% \begin{table}[t]
% 		\caption{Retrieval scores (\%) at both the object level (O-) and image level (I-) with different clustering numbers on the object retrieval validation set.}
%  		\vspace{-4mm}
%             \renewcommand\arraystretch{1.05}
% 		\begin{center}
% 		\resizebox{\linewidth}{!}{
% 			\begin{tabularx}{9.7cm}{c|cccc}
% 				\hline 
% 				 Cluster Number& O-Recall@1&O-mAP&I-Recall@1& I-mAP\\
% 				\hline
%                   50& 70.43&17.95&74.42&72.47\\
%                   100& \textbf{70.60}& \textbf{18.25} &\textbf{74.69}& \textbf{72.47}\\
%                   200& 69.98 & 17.75 & 74.11&71.99\\
%                   1000& 69.13&17.28&73.07&70.87\\
% 				\hline
% 			\end{tabularx}
% 			}
% 		\end{center}
% 		\vspace{-4mm}
% 		\label{tab:ab-cluster}
% 	\end{table}

(1) \textbf{Impact of the MS-GCEL group number.} We assessed the effect of the number of groups on the performance of the MS-GCEL method, with the cluster number fixed at 100. The results on the validation set are displayed in~\cref{tab:ab-group}, and they indicate that the best performance is attained when the number of groups is set to 4. Thus, in all of our experiments, we set the number of groups as 4.

(2) \textbf{Impact of the clustering number.} We examined the impact of the clustering number on deep feature embedding performance, with the number of groups fixed at 3. The results on the validation set are presented in \cref{tab:ab-cluster}, and it is evident that the best performance is achieved when the number of clusters is set to 100.

% (2) \textbf{Impact of the MS-GCEL method.} We assessed the influence of the MS-GCEL method on the object retrieval validation set. Results comparing its use with and without the MS-GCEL method are presented in~\cref{tab:ab-gcml}, demonstrating an improvement in performance when integrating the MS-GCEL method.
(3) \textbf{Impact of the MS-GCEL method.}
We assessed the influence of the PEL module and the CKD loss function in the proposed MS-GCEL approach on the object retrieval validation set. The ablation experiments are shown in~\cref{tab:ab-gcml}. The utilization of the PEL, which focuses on obtaining scale-specific information for feature encoding, enhances object-level retrieval performance. This indicates that the PEL assists in tackling the multi-scale challenge in object retrieval. The incorporation of the CKD loss function, which promotes information sharing and transfer among several PELs, significantly improves the model's retrieval performance.

\begin{table}
    \centering
    \caption{Retrieval scores (\%) at both the object level (O-) and image level (I-) with different hyperparameters on the object retrieval validation set.}
    \vspace{-4mm}
    \begin{subtable}{0.48\textwidth}
        \centering
        \caption{Different group numbers.}
        \vspace{-2mm}
        \resizebox{\linewidth}{!}{
			\begin{tabularx}{9.7cm}{c|cccc}
				\hline 
				 Group Number& O-Recall@1&O-mAP&I-Recall@1& I-mAP\\
				\hline
                  2& 70.16&17.74&74.05&72.57\\
                  3& 70.60& \textbf{18.25} &74.69& 72.47\\
                  4& \textbf{70.78} & \textbf{18.25} & \textbf{74.91}&\textbf{72.60}\\
                  5& 69.68&17.75&73.92&71.72\\
				\hline
			\end{tabularx}
			}
            % \vspace{-4mm}
            \label{tab:ab-group}
    \end{subtable}
    \hfill
    \begin{subtable}{0.48\textwidth}
        \centering
        \caption{Different clustering numbers}
        \vspace{-2mm}
        \resizebox{\linewidth}{!}{
			\begin{tabularx}{9.7cm}{c|cccc}
				\hline 
				 Cluster Number& O-Recall@1&O-mAP&I-Recall@1& I-mAP\\
				\hline
                  50& 70.43&17.95&74.42&72.47\\
                  100& \textbf{70.60}& \textbf{18.25} &\textbf{74.69}& \textbf{72.47}\\
                  200& 69.98 & 17.75 & 74.11&71.99\\
                  1000& 69.13&17.28&73.07&70.87\\
				\hline
			\end{tabularx}
			}
            \vspace{-2mm}
            \label{tab:ab-cluster}
    \end{subtable}
\end{table}

 % \begin{table}[t]
	% 	\caption{Retrieval scores (\%) at both the object level (O-) and image level (I-) with different group numbers on the object retrieval validation set.}
 % 		% \vspace{-4mm}
 %            \renewcommand\arraystretch{1.05}
	% 	\begin{center}
	% 	\resizebox{0.5\linewidth}{!}{
	% 		\begin{tabularx}{9.7cm}{c|cccc}
	% 			\hline 
	% 			 Group Number& O-Recall@1&O-mAP&I-Recall@1& I-mAP\\
	% 			\hline
 %                  2& 70.16&17.74&74.05&72.57\\
 %                  3& 70.60& \textbf{18.25} &74.69& 72.47\\
 %                  4& \textbf{70.78} & \textbf{18.25} & \textbf{74.91}&\textbf{72.60}\\
 %                  5& 69.68&17.75&73.92&71.72\\
	% 			\hline
	% 		\end{tabularx}
	% 		}
	% 	\end{center}
	% 	% \vspace{-2mm}
	% 	\label{tab:ab-group}
	% \end{table}

\begin{table}[!t]
    \centering
    \caption{The ablation experiments of PEL and CKD.}
    \vspace{-4mm}
    \small % Adjusts font size to fit the table on the page
     \resizebox{0.5\textwidth}{!}{
    \begin{tabular}{l|c c c c}
        % \toprule
        \hline
        {Methods} & {O-Recall@1} & {O-mAP} & {I-Recall@1} & {I-mAP}\\
        % \midrule
        \hline
        STML & 69.12 & 16.79 & 73.07 & 70.19 \\
        STML+PEL    & 69.29 & 17.62 & 72.98 & 71.13 \\
        STML+PEL+CKD & \textbf{70.78} & \textbf{18.25} & \textbf{74.91}&\textbf{72.60}  \\
        % \bottomrule
        \hline
    \end{tabular}
    }
    % \vspace{-2mm}
    \vspace{-4mm}
    \label{tab:ab-gcml}
\end{table}

\subsection{More Retrieval Examples}
\label{sec: more_examples}
\begin{figure*}[!t]
    \begin{center}
    \includegraphics[width=0.9\linewidth]{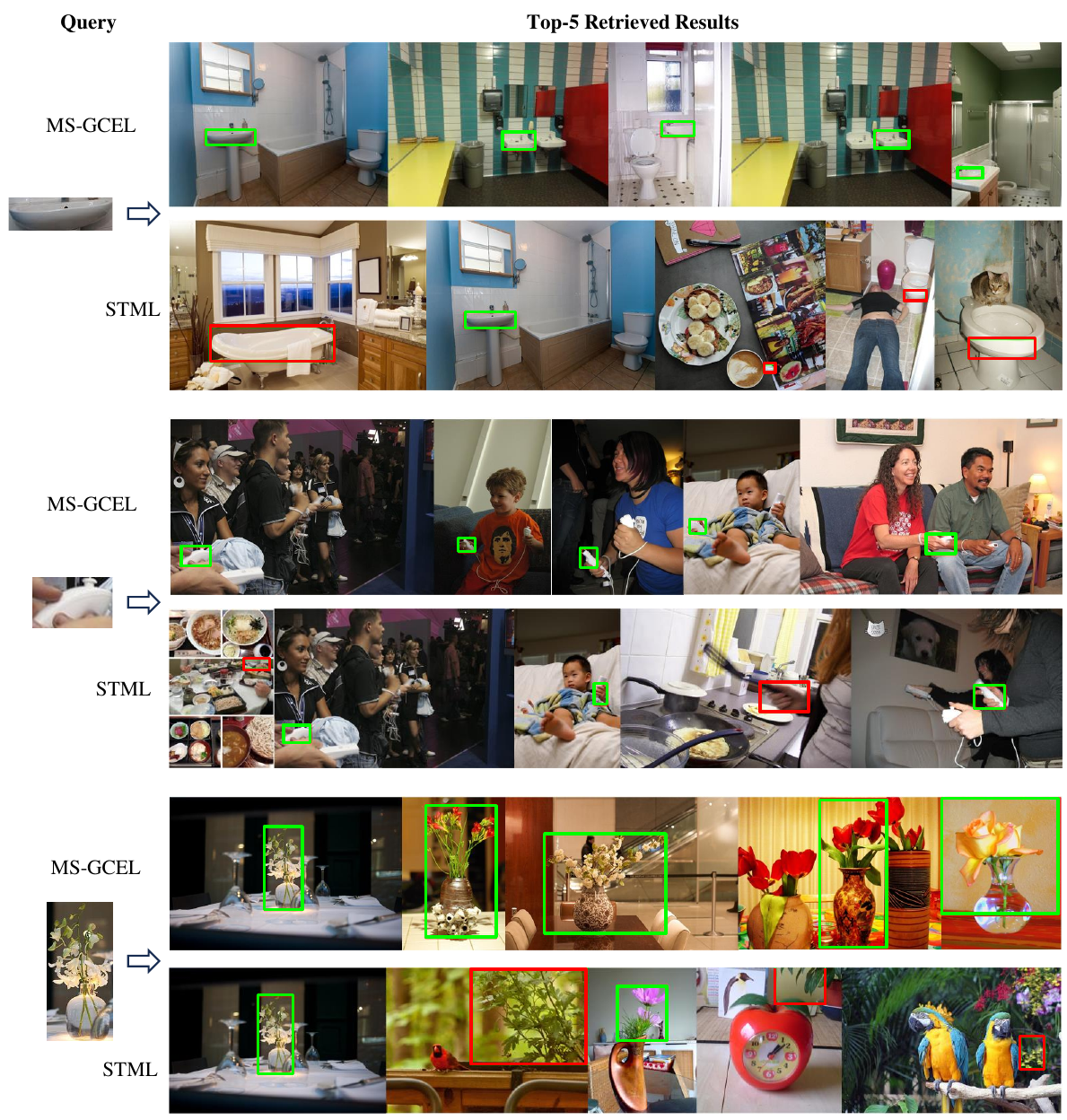}
    \end{center}
    \vspace{-4mm}
	\caption{The comparison of three object retrieval examples between the MS-GCEL and STML\cite{kim2022self} methods. Retrieved objects with \textcolor{green}{green} boxes are correct ones with the same class label as the query object. Those with \textcolor{red}{red} boxes are incorrect results.}
	\vspace{-8mm}
	\label{fig:compare_case.png}
\end{figure*}

\cref{fig:compare_case.png} illustrates the comparison of three object retrieval examples between the MS-GCEL and STML\cite{kim2022self} methods. The first retrieval example is based on the GoogLeNet backbone network, while the second and third retrieval examples are based on the MiT-B2 backbone network. These examples demonstrate that the MS-GCEL method is capable of returning more relevant objects associated with the query object.

\begin{figure*}[!t]
    \begin{center}
    \includegraphics[width=0.9\linewidth]{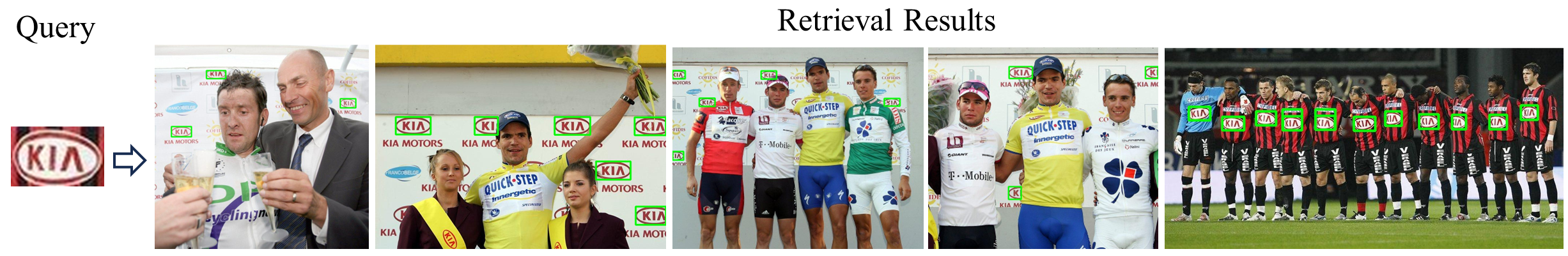}
    \end{center}
    \vspace{-4mm}
	\caption{Logo retrieval example on the BelgaLogos using the proposed MS-GCEL method.}
	\vspace{-4mm}
	\label{fig:retrieval-examples}
\end{figure*}
~\cref{fig:retrieval-examples} provides logo retrieval example showcasing the effectiveness of the proposed MS-GCEL method. These examples illustrate the ability to ground multiple objects within the retrieved images.

\begin{figure*}[t]
    \begin{center}
    \includegraphics[width=0.9\linewidth]{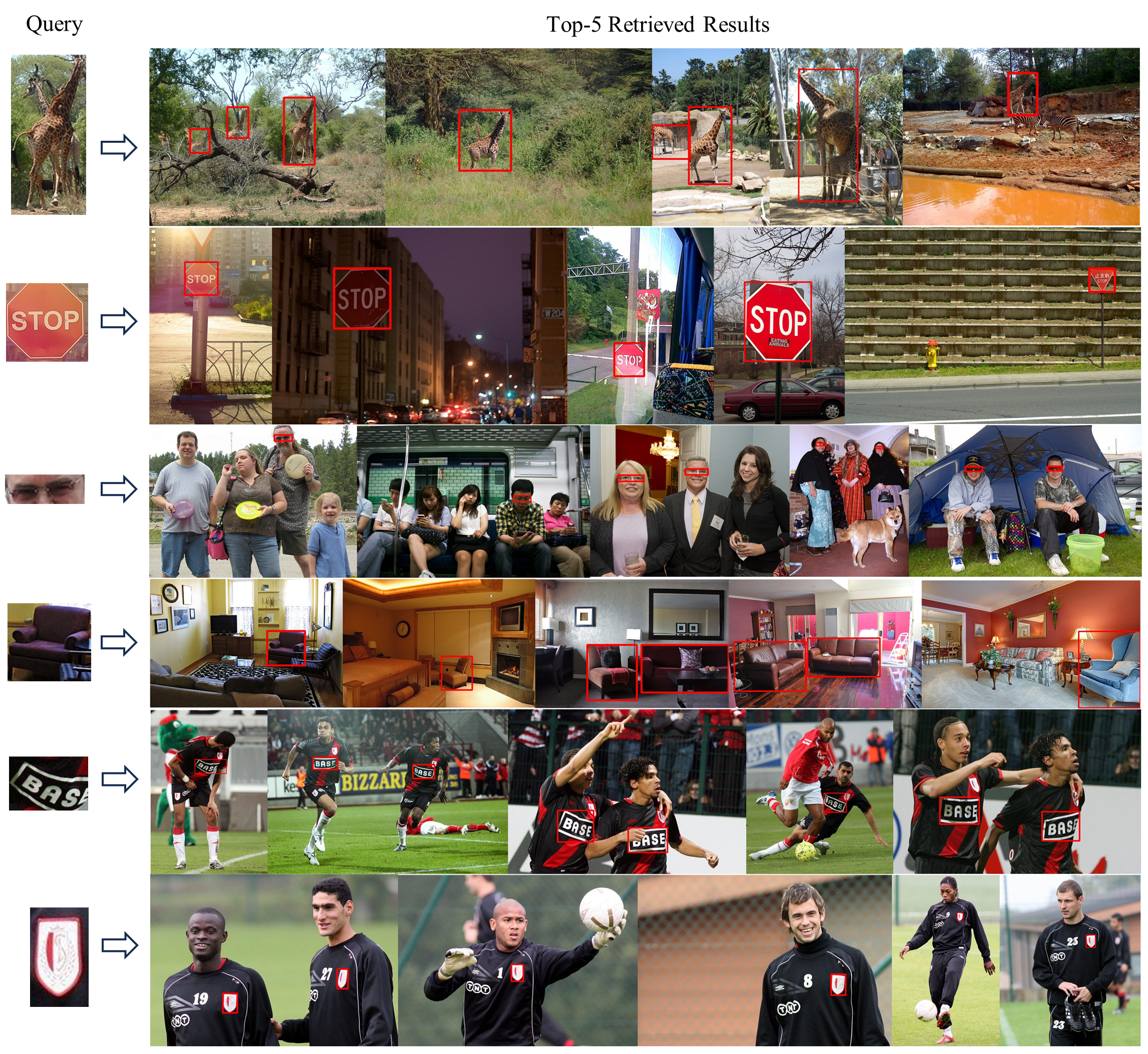}
    \end{center}
    \vspace{-4mm}
	\caption{Six object retrieval examples using the fine-tuned model based on the MS-GCEL method with parameters initialized using MoCo-V3.}
	\vspace{-4mm}
	\label{fig:supplemental_more_case}
\end{figure*}

In \cref{fig:supplemental_more_case}, we present six retrieval examples. These examples were retrieved using the fine-tuned model based on the MS-GCEL method with parameters initialized using MoCo-V3 \cite{chen2021}. During retrieval, the top five images containing the query object were retrieved. Subsequently, object localization was conducted in these retrieved images using a similarity score (feature distance).

\begin{figure*}[t]
    \begin{center}
    \includegraphics[width=0.9\linewidth]{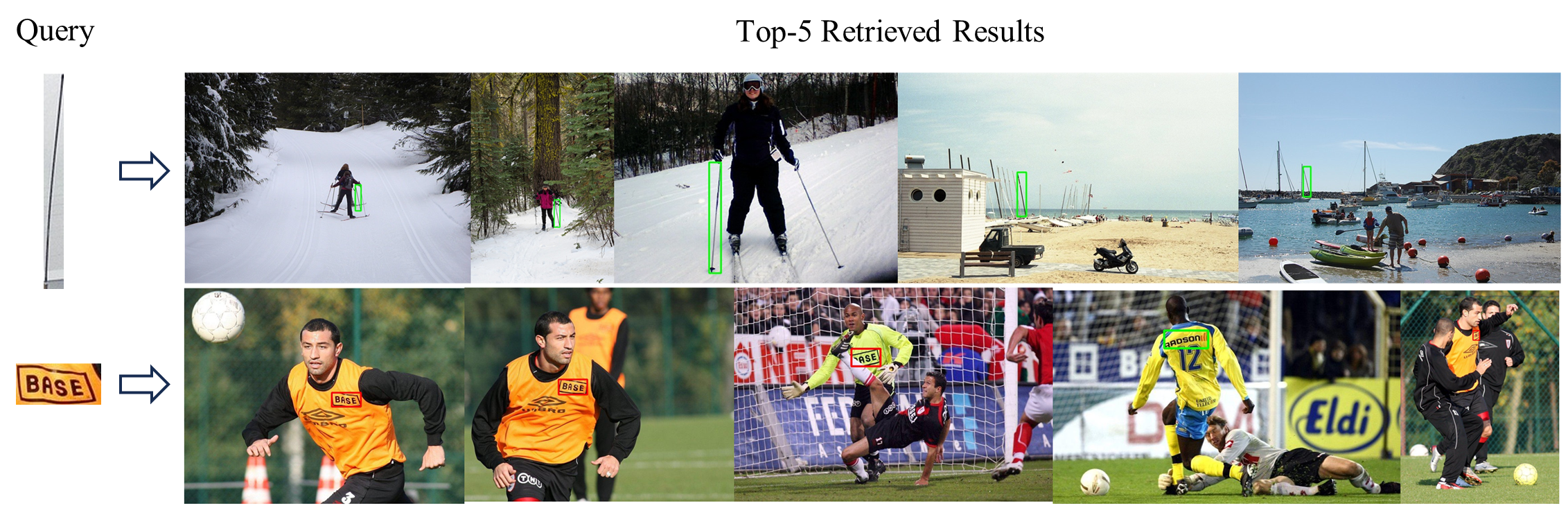}
    \end{center}
    \vspace{-4mm}
	\caption{Two failure cases.  The first line of the query image is antenna, but the search returns ski-pole and mast. The image retrieved in the second row is the BASE logo. However, the fourth image is not the BASE logo, but other text patterns.}
	\vspace{-4mm}
	\label{fig:failure_case.png}
\end{figure*}

\cref{fig:failure_case.png} shows two failure cases,  the first line of the query image in the \cref{fig:failure_case.png} is antenna, but the search returns ski-pole and mast. This is because antenna, ski-pole and mast are essentially the same in shape, and distinguishing between them requires more contextual semantic information. The image retrieved in the second row is the BASE logo. However, the fourth image is not the BASE logo, but other text patterns.
 
\begin{table*}[t]
		\caption{Category splits for the open-vocabulary scenario are provided. Known classes are highlighted in light yellow, while unknown classes are highlighted in light blue.}
 		\vspace{-4mm}
		\begin{center}
		\resizebox{\linewidth}{!}{
			\begin{tabularx}{21.5cm}{ccccccccccc}
              \hline
                \rowcolor{LightGrey}\multicolumn{11}{c}{Task1}\\
                    \hline
                \rowcolor{OldLace} &person &bicycle &car &motorcycle &airplane &bus &train &boat &bird &cat\\
                 \rowcolor{OldLace} &dog &horse &sheep &cow &bottle &chair &couch &potted plant &dining table &tv\\
                  \rowcolor{AliceBlue}&truck &traffic light &fire hydrant &stop sign &parking meter &bench &elephant &bear &zebra &giraffe\\
                \rowcolor{AliceBlue}&backpack &umbrella &handbag &tie &suitcase &microwave &oven &toaster &sink &refrigerator\\

                \rowcolor{AliceBlue}&frisbee &skis &snowboard &sports ball &kite &baseball bat &baseball glove &skateboard &surfboard &tennis racket\\
                \rowcolor{AliceBlue}&banana &apple &sandwich &orange &broccoli &carrot &hot dog &pizza &donut &cake\\

                \rowcolor{AliceBlue}&bed &toilet &laptop &mouse &remote &keyboard &cell phone &book &clock &vase\\ 
                \rowcolor{AliceBlue}&scissors &teddy bear &hair drier &toothbrush &wine glass &cup &fork &knife &spoon &bowl\\
                    \hline
			\end{tabularx}
			}
   
            \resizebox{\linewidth}{!}{
			\begin{tabularx}{21.5cm}{ccccccccccc}
              \hline
                \rowcolor{LightGrey}\multicolumn{11}{c}{Task2}\\
                    \hline
                \rowcolor{OldLace} &person &bicycle &car &motorcycle &airplane &bus &train &boat &bird &cat\\
                 \rowcolor{OldLace} &dog &horse &sheep &cow &bottle &chair &couch &potted plant &dining table &tv\\
                  \rowcolor{OldLace}&truck &traffic light &fire hydrant &stop sign &parking meter &bench &elephant &bear &zebra &giraffe\\
                \rowcolor{OldLace}&backpack &umbrella &handbag &tie &suitcase &microwave &oven &toaster &sink &refrigerator\\

                \rowcolor{AliceBlue}&frisbee &skis &snowboard &sports ball &kite &baseball bat &baseball glove &skateboard &surfboard &tennis racket\\
                \rowcolor{AliceBlue}&banana &apple &sandwich &orange &broccoli &carrot &hot dog &pizza &donut &cake\\

                \rowcolor{AliceBlue}&bed &toilet &laptop &mouse &remote &keyboard &cell phone &book &clock &vase\\ 
                \rowcolor{AliceBlue}&scissors &teddy bear &hair drier &toothbrush &wine glass &cup &fork &knife &spoon &bowl\\
                    \hline
			\end{tabularx}
			}

                \resizebox{\linewidth}{!}{
			\begin{tabularx}{21.5cm}{ccccccccccc}
              \hline
                \rowcolor{LightGrey}\multicolumn{11}{c}{Task3}\\
                    \hline
                \rowcolor{OldLace} &person &bicycle &car &motorcycle &airplane &bus &train &boat &bird &cat\\
                 \rowcolor{OldLace} &dog &horse &sheep &cow &bottle &chair &couch &potted plant &dining table &tv\\
                  \rowcolor{OldLace}&truck &traffic light &fire hydrant &stop sign &parking meter &bench &elephant &bear &zebra &giraffe\\
                \rowcolor{OldLace}&backpack &umbrella &handbag &tie &suitcase &microwave &oven &toaster &sink &refrigerator\\

                \rowcolor{OldLace}&frisbee &skis &snowboard &sports ball &kite &baseball bat &baseball glove &skateboard &surfboard &tennis racket\\
                \rowcolor{OldLace}&banana &apple &sandwich &orange &broccoli &carrot &hot dog &pizza &donut &cake\\

                \rowcolor{AliceBlue}&bed &toilet &laptop &mouse &remote &keyboard &cell phone &book &clock &vase\\ 
                \rowcolor{AliceBlue}&scissors &teddy bear &hair drier &toothbrush &wine glass &cup &fork &knife &spoon &bowl\\
                    \hline
			\end{tabularx}
			}

                \resizebox{\linewidth}{!}{
			\begin{tabularx}{21.5cm}{ccccccccccc}
              \hline
                \rowcolor{LightGrey}\multicolumn{11}{c}{Task4}\\
                    \hline
                \rowcolor{OldLace} &person &bicycle &car &motorcycle &airplane &bus &train &boat &bird &cat\\
                 \rowcolor{OldLace} &dog &horse &sheep &cow &bottle &chair &couch &potted plant &dining table &tv\\
                  \rowcolor{OldLace}&truck &traffic light &fire hydrant &stop sign &parking meter &bench &elephant &bear &zebra &giraffe\\
                \rowcolor{OldLace}&backpack &umbrella &handbag &tie &suitcase &microwave &oven &toaster &sink &refrigerator\\

                \rowcolor{OldLace}&frisbee &skis &snowboard &sports ball &kite &baseball bat &baseball glove &skateboard &surfboard &tennis racket\\
                \rowcolor{OldLace}&banana &apple &sandwich &orange &broccoli &carrot &hot dog &pizza &donut &cake\\

                \rowcolor{OldLace}&bed &toilet &laptop &mouse &remote &keyboard &cell phone &book &clock &vase\\ 
                \rowcolor{OldLace}&scissors &teddy bear &hair drier &toothbrush &wine glass &cup &fork &knife &spoon &bowl\\
                    \hline
			\end{tabularx}
			}
		\end{center}
		\vspace{-4mm}
		\label{tab:tasksplit}
	\end{table*}
\section{Conclusion}
In this paper, we have addressed the challenges associated with general object retrieval for OK-VQA by introducing the multi-scale group collaborative embedding learning (MS-GCEL) method. We have established a comprehensive training and evaluation pipeline using objects extracted from the COCO 2017 and VOC 2007 datasets. Our experiments encompass different backbone networks and initialization methods, with evaluations conducted on BelgaLogos, Visual Genome, and LVIS datasets, and one curated OK-VQA dataset.
The results demonstrated the effectiveness of the proposed MS-GCEL method in enhancing object retrieval performance in OK-VQA scenario. Nevertheless, there remains substantial room for improvement in this direction, as the target objects typically constitute a relatively small and unpredictable portion of an image, thereby impacting the overall object retrieval performance.

\section*{Acknowledgments}
 This work was supported in part by the National Natural Science Foundation of China under Grant 62202499 and 62463002, in part by the State Key Program of National Natural Science Foundation of China under Grant 62233018, in part by the Beijing Natural Science Foundation under grant L231012. We are grateful to  the High Performance Computing Center of Central South University for partial support of this work.

% \input{sec/X3_appendix}




{
    % \small
    \bibliographystyle{elsarticle-num}
    \bibliography{main}
}
\end{document}